\documentclass{article} % For LaTeX2e
\usepackage{iclr2023_conference,times}

\usepackage{amsmath,amsfonts,bm}

\def\eqref#1{equation~\ref{#1}}
\def\1{\bm{1}}

\DeclareMathAlphabet{\mathsfit}{\encodingdefault}{\sfdefault}{m}{sl}
\SetMathAlphabet{\mathsfit}{bold}{\encodingdefault}{\sfdefault}{bx}{n}

\usepackage{hyperref}
\usepackage[utf8]{inputenc} % allow utf-8 input
\usepackage[T1]{fontenc}    % use 8-bit T1 fonts
\usepackage{hyperref}       % hyperlinks
\usepackage{url}            % simple URL typesetting
\usepackage{booktabs}       % professional-quality tables
\usepackage{amsfonts}       % blackboard math symbols
\usepackage{nicefrac}       % compact symbols for 1/2, etc.
\usepackage{microtype}      % microtypography
\usepackage{xcolor}         % colors \LaTeX\ Companio
\usepackage{wrapfig}
\usepackage{color}
\usepackage{adjustbox}
\usepackage{graphics}
\usepackage{amsmath}
\usepackage{enumitem}
\usepackage[ruled,linesnumbered,vlined]{algorithm2e}
\usepackage{algorithmicx}
\usepackage{algpseudocode}
\usepackage[english]{babel}
\newtheorem{theorem}{Theorem}

\usepackage{mathabx}
\usepackage{multirow}
\usepackage[normalem]{ulem}
\useunder{\uline}{\ul}{}

\definecolor{mydarkblue}{rgb}{0,0.08,0.45}
\definecolor{myblue}{HTML}{3b75c3}
\definecolor{myred}{HTML}{E33222}
\definecolor{mygreen}{HTML}{438773}
\definecolor{mymaroon}{RGB}{142,27,19}
\definecolor{maroon}{HTML}{800000}
\definecolor{mycite}{cmyk}{0.55,1,0,0.15}
\definecolor{codeblue}{rgb}{0.25,0.5,0.5}
\definecolor{codekw}{rgb}{0.85, 0.18, 0.50}
\definecolor{codegreen}{rgb}{0,0.6,0}
\definecolor{codegray}{rgb}{0.5,0.5,0.5}
\definecolor{codepurple}{rgb}{0.58,0,0.82}
\definecolor{backcolour}{rgb}{0.95,0.95,0.92}
\hypersetup{
    colorlinks=true,
    citecolor=blue,
    linkcolor=myblue,
    urlcolor=maroon
          }

\title{Graph Contrastive Learning with \\ Cross-view Reconstruction}

\author{Qianlong Wen$^{1}$, Zhongyu Ouyang$^{1}$, Chunhui Zhang$^{2}$, Yiyue Qian$^{1}$, \\ \bf Yanfang Ye$^{1}$, Chuxu Zhang$^{2}$ \\
$^1$University of Notre Dame, $^2$Brandeis University \\
{$^1$\tt $\{$qwen,zouyang2,yqian5,yye7$\}$@nd.edu};\\
{$^2$\tt $\{$chunhuizhang,chuxuzhang$\}$@brandeis.edu}
}

\iclrfinalcopy % Uncomment for camera-ready version, but NOT for submission.
\begin{document}

\maketitle

\begin{abstract}
Graph self-supervised learning is commonly taken as an effective framework to tackle the supervision shortage issue in the graph learning task. 
% and our-of-distribution generalization issue in graph learning task.  
% Although different graph self-supervised learning strategies have been proposed to tackle the supervision shortage issue in graph learning tasks, 
Among different existing graph self-supervised learning strategies, graph contrastive learning (GCL) has been one of the most prevalent approaches to this problem. Despite the remarkable performance those GCL methods have achieved, existing GCL methods that heavily depend on various manually designed augmentation techniques still struggle to alleviate the feature suppression issue without risking losing task-relevant information.  Consequently, the learned representation is either brittle or unilluminating. In light of this, we introduce the \textbf{Graph} Contrastive Learning with \textbf{C}ross-\textbf{V}iew Reconstruction (GraphCV), which follows the information bottleneck principle to learn minimal yet sufficient representation from graph data. Specifically, GraphCV aims to elicit the predictive (useful for downstream instance discrimination) and other non-predictive features separately. Except for the conventional contrastive loss which guarantees the consistency and sufficiency of the representation across different augmentation views, we introduce a cross-view reconstruction mechanism to pursue the disentanglement of the two learned representations. Besides, an adversarial view perturbed from the original view is added as the third view for the contrastive loss to guarantee the intactness of the global semantics and improve the representation robustness. We empirically demonstrate that our proposed model outperforms the state-of-the-art on graph classification task over multiple benchmark datasets.

\end{abstract}

% \vspace{-0.1in}
\section{Introduction}
\label{sec: intro}
\vspace{-0.1in}
Graph representation learning (GRL) has attracted significant attention due to its widespread applications in the real-world interaction systems, such as social, molecules, biological and citation networks \citep{GRL_survey}. The current state-of-the-art supervised GRL methods are mostly based on Graph Neural Networks (GNNs) \citep{GCN, GAT, hamilton_inductive_2017, GIN}, which require a large amount of task-specific supervised information. Despite the remarkable performance, they are usually limited by the deficiency of label supervision in real-world graph data due to the fact that it is usually easy to collect unlabeled graph but very costly to obtain enough annotated labels, especially in certain fields like biochemistry. 
Therefore, many recent works \citep{GCC, ContrastMultiView, InfoGraph} study how to fully utilize the unlabeled information on graph and further stimulate the application of self-supervised learning (SSL) for GRL where only limited or even no label is needed. 

\begin{figure*}[t]
  \centering
  \vspace{-0.4in}
  \includegraphics[width=0.9\linewidth]{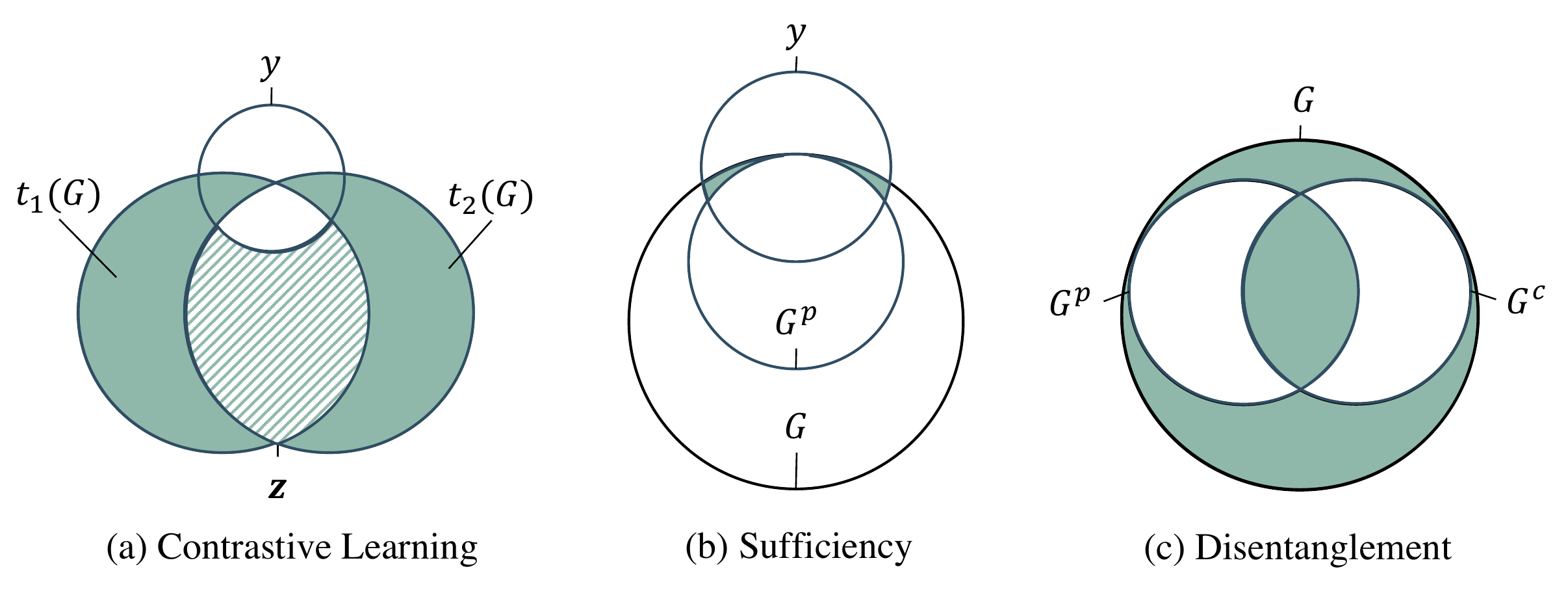}
    \vspace{-0.2in}
  \caption{Illustration of the relation between graph $G$, label $y$, predictive feature subsets $G^{p}$ and non-predictive feature subset $G^{c}$ in terms of information entropy. Ideally, the green areas in the three figures is null.
  (a) The usual optimization result of graph contrastive learning, where the shared features of two augmentation view is extracted for the learned representation $\mathbf{z}$. Owing to the lack of supervision or domain knowledge, redundant and biased information (shadow area) is usually included in $\mathbf{z}$; (b) $G^{p}$ cover the feature subset which is sufficient to make correct graph label identification ($I(y; G \mid G^{p})=0$), other features ($G^{c}$) is either useless or misguiding; (c) $G^{p}$ and $G^{c}$ are supposed to be mutually disentangled with each other ($I(G^{p}; G^{c})=0$).
%   because $G^{p}$ only cover the rational feature subset for label identification while $G^{c}$ does not contain any rational features. 
  The union of them cover all the features of original data.}
  \vspace{-0.35in}
  \label{fig: mutual_info}
\end{figure*}

As a prevalent and effective strategy of SSL, contrastive learning follows the mutual information maximization principle (InfoMax) \citep{DGI} to maximize the agreements of the positive pairs while minimizing that of the negative pairs in the embedding space. 
% In particular, graph contrastive learning (GCL) methods \citep{GCC, ContrastMultiView, GraphCL} usually take two different augmentation views from the same graph as the positive pair and maintain a high consistency between the learned representations of the views to preserve the invariance of graphs properties while ignore the nuance. 
% To be more specific, each commonly-used augmentation technique $A(\cdot)$ transforms the original graph data $G$ in a certain way and therefore add random perturbation on original graph $G$. Then, the GCL paradigm aims to train the graph feature encoder $f(\cdot)$ to achieve the invariance goal, i.e., $f(A(G))=f(G)$. 
However, the graph contrastive learning paradigm guided by the InfoMax principle is insufficient to learn robust and transferable representation. State-of-the-art GCL methods \citep{GCC, ContrastMultiView, GraphCL} usually rely on augmentor(s) $t(\cdot)$ (e.g., Identity, Subgraph Sampling, Node Dropping, Edge Removing and Attributes Masking.) applied to the anchor graph $G$ to generate a positive pair of graphs $t_{1}(G)$ and $t_{2}(G)$. 
Then, the graph feature encoder $f$ will be trained to ensure the representation consistency within the positive pair, i.e., $\mathbf{z}=f(t_{1}(G))=f(t_{2}(G))$. Consequently, such training strategy is heavily dependent on the choice and strength of graph augmentation techniques. 
To be more specific, moderate graph augmentation will push encoders to capture redundant and biased information \citep{on_mutual_info}, which could inadvertently suppress the space of important predictive features and negatively affect the representation transferability via the so-called "shortcut" solution \citep{shortcut, automatic_shortcut}.  
A more intuitively illustration is provided in the Figure \ref{fig: mutual_info} (a), where the shared part of the two augmentation view include both predictive information (the overlapping area with $y$) and non-predictive information (shadow area).
Such optimization result usually yield lower contrastive loss, however, it has been empirically proved that the redundant information could lead to poor robustness~\citep{CL_shortcut}, especially under the out-of-distribution (OOD) setting~\citep{ood_general}. We provide a showcase example in Appendix \ref{appendix: ood_case} to illustrate the OOD scenario on graph learning task. 
On the other hand, overly aggressive augmentation may easily lead to another extreme where many predictive features are randomly dropped and the learned representation does not contain sufficient predictive information for downstream instance discrimination.
Recent works \citep{AD-GCL, RGCL, JOAO} propose to use automated augmentations to extract the invariant rationale features \citep{DIR, EERM}. These methods assume the most salient sub-structure (those are resistant to graph augmentation) is sufficient to make rational and correct label identification, and thereby implement trainable augmentation operations (e.g., edge deleting, node dropping) to strictly regularize the graph topological structure. Despite that these methods can alleviate the aforementioned feature suppression issue to some extent, they still suffer from inherent limitation. The harsh regularization may force the encoders focusing on the easy-learned "shallower" features (e.g. graph size and node degree), which might be helpful under certain domains but 
% are unable to cover others~\citep{size_ood}, thus still fail to guarantee higher robustness. 
not necessarily for others~\citep{size_ood}, thus fail to guarantee stronger robustness.
%% Figure 1 showcase the ...
Therefore, the GCL methods guided with the saliency philosophy is not flexible enough to balance the representation sufficiency and robustness without the guidance of explicit domain knowledge. 
To reconcile the robustness and sufficiency of the learned representation, a method which can reduce redundant and biased information without sacrificing the sufficiency of the predictive graph features is in urgent need.

Recently, the information bottleneck (IB) principle~\citep{IB} has been introduced to the graph learning,
% where minimal yet sufficient information is extracted for representation learning. 
which encourages extracting minimal yet sufficient information for representation learning.
The core idea of IB principle is in accordance with the ultimate optimization objective to solve the feature suppression issue \citep{CL_shortcut}, thus shed more light on this problem.  
Moreover, the representation learning guided by the IB principle has been empirically proved to generate more robust and transferable representations at different domains~\citep{GIB}. 
Therefore, a graph contrastive learning framework in accordance with the IB principle is promising in balancing the representation robustness and sufficiency.
% Given an input graph $G$, we use $G^{p}$ and $G^{c}$ to denote its predictive feature subset and the complementary non-predictive feature subset. 
Given an input graph $G$, we denote $G^{p}$ and $G^{c}$ as its predictive feature subset and the complementary non-predictive feature subset, respectively.
According to the assumption of recent studies about rationale invariance discover~\citep{DIR, EERM}, the two features subsets would the satisfy $I(y; G \mid G^{p})=0$ (sufficiency condition) and disentanglement condition (i.e., $I(G^{p}; G^{c})=0$). We illustrate the relations among the two feature subsets and $G$ in Figure \ref{fig: mutual_info} (b) and (c).
It is inevitable that the learned representation maintains some redundant information for a specific downstream task. However, a GCL framework under the guidance of the IB principle is expected to suppress the feature space of $G^{c}$ as much as possible while keeping the predictive feature $G^{p}$ intact simultaneously in the learned representation. 

In this paper, we propose the novel \textbf{Graph} Contrastive Learning with \textbf{C}ross-\textbf{V}iew Reconstruction, named GraphCV, to pursue the optimization objective of the IB principle. Specifically, GraphCV consists of a graph encoder followed with two decoders that are trained to extract information specific to the predictive and non-predictive features, respectively. 
To approximate the disentanglement objective, we propose the reconstruction-based representation learning scheme, including the intra-view and inter-view reconstructions, to reconstruct the original learned representation with the two separated feature subsets. 
% Meanwhile, an adversarial graph perturbed from original view is added as the third view of the contrastive loss besides the predictive relevant representations of the two augmented graph views to further improve the representation robustness and prevent it form collapsing into partial or even trivial features.
Furthermore, the encoded representation from the original view perturbed in the adversarial fashion serves as the third view when computing the contrastive loss, apart from the predictive relevant representations of the two augmentation views, to further improve the representations' robustness and prevent them from collapsing into partial or even trivial ones.
We provide theoretical analysis to show that GraphCV is capable to learn minimal sufficient representations.
% with the designs above. 
Finally, we conduct experiments to validate the effectiveness of GraphCV over the commonly-used graph benchmark datasets. The experimental results demonstrate that GraphCV achieves significant performance gains over different datasets and settings compared with state-of-the-art baselines. 

The main contributions of this work are summarized from three aspects: (i) We propose the GraphCV to alleviate the feature suppression issue with the cross-view reconstruction mechanism; (ii) We provide solid theoretical analysis on our model designs; (iii) Thorough experiments are conducted to demonstrate the robustness and transferability of the learned representations via GraphCV.
% that GraphCV can mitigate the "shortcut" solution in contrastive learning and significantly outperforms the state-of-the-art baselines over multiple graph classification benchmark datasets. 

\vspace{-0.2in}
\section{Preliminaries}

\vspace{-0.1in}
\subsection{Graph Representation Learning}
\vspace{-0.1in}
\label{sec: GRL}
In this work, we focus on the graph-level task, let $\mathcal{G}=\left\{G_{i} = (V_{i}, E_{i})\right\}_{i=1}^{N}$ denote a graph dataset with N graphs, where $V_{i}$ and $E_{i}$ are the node set and edge set of graph $G_{i}$, respectively. We use $x_{v} \in \mathbb{R}^{d}$ and $x_{e} \in \mathbb{R}^{d}$ to denote the attribute vector of each node $v \in V_{i}$ and edge $e \in E_{i}$. Each graph is associated with a label, denoted as $y_{i}$, the goal the graph representation learning is to learn an encoder $f: G_{i} \rightarrow \mathbb{R}^{d}$ so that the learned representation $\mathbf{z}_{i} = f(G_{i})$ is sufficient to predict $y_{i}$ related to the downstream task. We clarify the sufficiency of $\mathbf{z}_{i}$ as containing 
% the same amount of information as $G_{i}$ for label identification \citep{sufficiency}, 
no less information of the label of $G_{i}$ \citep{sufficiency}, 
and it is formulated as: 
\begin{equation}
    I(G_{i} ; y_{i} \mid \mathbf{z}_{i})=0,
    \label{eq: info_sufficiency}
\end{equation}
where $I\left( ; \right)$ denotes the mutual information between two variables. 
% We demonstrate the general optimization result of classical representation learning in Figure~\ref{fig: intro}(a).

\vspace{-0.1in}
\subsection{Contrastive Learning}
\vspace{-0.1in}
\label{sec: CL}
Contrastive Learning (CL) is a self-supervised representation learning method which leverages instance-level identity for supervision. 
% It follows the InfoMax principle to push the learned representations agree with each other under proper transformations. 
During the training phase, each graph $G$ firstly goes through proper data augmentation to generate two data augmentation views $t_{1}(G)$ and $t_{2}(G)$, where $t_{1}(\cdot)$ and $t_{2}(\cdot)$ are two augmentation operators. Then, the CL method encourages the encoder $f$ (a backbone network plus a projection layer) to map $t_{1}(x)$ and $t_{2}(x)$ closer in the hidden space so that the learned representations $\mathbf{z}_{1}$ and $\mathbf{z}_{2}$ maintain all the information shared by $t_{1}(G)$ and $t_{2}(G)$. The learning of the encoder is usually directed by a contrastive loss, such as NCE loss \citep{NCE}, InfoNCE loss \citep{InfoNCE} and NT-Xent loss \citep{SimCLR}. 
In Graph Contrastive Learning (GCL), we usually adopt a GNN, such as GCN \citep{GCN} or GIN \citep{GIN}, as the backbone network, and the commonly-used graph data augmentation operators~\citep{GraphCL}, such as node dropping, edge perturbation, subgraph sampling, and attribute masking. 

All the GCL-based methods are built on the assumption that augmentations do not 
% change the information regarding to the label. 
break the sufficiency requirement to make correct prediction. 
Here, we follow \citep{robust_rep} to clear up the definition of mutual redundancy. $t_{1}(G)$ is redundant to $t_{2}(G)$ with respect of $y$ iff $t_{1}(G)$ and $t_{2}(G)$ share the same predictive information. 
% In CL, $t_{1}(G)$ and $t_{2}(G)$ are supposed to be mutually redundant, 
Mathematically, the mutual redundancy in CL exists when: 
\begin{equation}
    I(t_{1}(G) ; y \mid t_{2}(G)) = I(t_{2}(G) ; y \mid t_{1}(G)) = 0.
    \label{eq: info_redundancy}
\end{equation}
Although GCL-based methods are usaully capable to extract useful information for label identification, it is unavoidable to include non-predictive features under the SSL setting owing lack of explicit domain knowledge. There exist the situation (e.g., OOD setting) that the latent space of learned representation is dominated by non-predictive features in SSL ~\citep{CL_FS} and it is no more informative enough to make correct prediction. 
Therefore, feature suppression is not just a prevalent issue in supervised learning, but also in SSL. Due to the page limitation, we provide more discussion about the relation between feature suppression and GCL in Appendix \ref{appendix: feature_suppression}

\section{Proposed Model}
\vspace{-0.1in}

\begin{figure*}[t]
  \centering
  \vspace{-0.1in}
  \includegraphics[width=\linewidth]{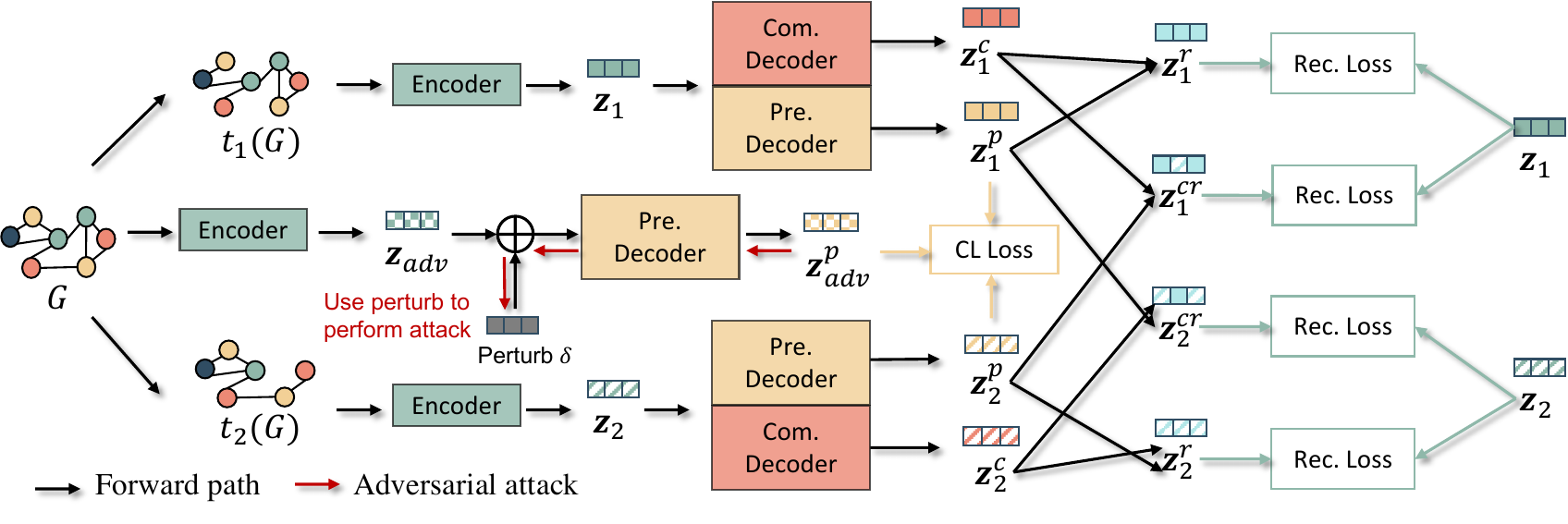}
  \vspace{-0.25in}
  \caption{The illustration of the proposed GraphCV. (1) Graph augmentations are applied to the input graph $G$ to produce two augmented graphs, which are then fed into the shared graph encoder $f(\cdot)$ to generate two graph representations $\mathbf{z}_{1}$ and $\mathbf{z}_{2}$. (2) $\mathbf{z}_{1}$ and $\mathbf{z}_{2}$ are used as the inputs of the two decoder to generate two pairs of graph representations, $\mathbf{z}^{p}$ captures the predictive factors and $\mathbf{z}^{c}$ keep other complementary non-predictive features. Then we use the two pairs of representations to reconstruct $\mathbf{z}_{1}$ and $\mathbf{z}_{2}$ in both of the intra-view and inter-view. (3) An adversarial sample generated by $G$ will go through the same procedure to generate $\mathbf{z}_{adv}^{p}$. We take it as the third view besides $\mathbf{z}_{1}^{p}$ and $\mathbf{z}_{2}^{p}$ in CL guarantee the $\mathbf{z}^{p}$ can keep the global semantics.}
    \vspace{-0.1in}
  \label{fig: model}
\end{figure*}

In this section, we introduce the details of our proposed GraphCV whose framework is shown in Figure \ref{fig: model}. Corresponding theoretical analysis are provided to justify the rationality of our designs. 
Before diving into the details of GraphCV, we briefly introduce the overall framework of our model. 

The proposed GraphCV model is designed in accordance with the IB principle to extract minimal yet sufficient representation through the designed cross-view reconstruction mechanism. 
Given $f(\cdot)$ as the graph encoder, we aim to map the graph representation $\mathbf{z}=f(G) \in \mathbb{R}^{d}$ into two different feature spaces
% $\mathbf{z} = (\mathbf{z}^{p}, \mathbf{z}^{c})$, 
$(\mathbf{z}^{p}, \mathbf{z}^{c})$, 
where $\mathbf{z}^{p} \in \mathbb{R}^{d}$ is expected to be specific to the predictive information $G^{p}$, and $\mathbf{z}^{c} \in \mathbb{R}^{d}$ is optimized to elicit the complementary non-predictive factors $G^{c}$. 
% Here, we follow the foundational assumption of GCL framework that mild augmentation will not break the mutual redundancy of two augmentation views for label \citep{SimCLR, hard_negative} and thus push $\mathbf{z}^{p}$ to inherent the invariant part of two augmentation views. Meanwhile, those features that are sensitive to augmentation will be taken as non-predictive factors and stored in $\mathbf{z}^{c}$.
Later, we reconstruct the representation $\mathbf{z}$ with the feature subsets mapped from same and different augmentation views to approximate the disentanglement objective demonstrated in Figure \ref{fig: mutual_info}.
By separating the learned representation into two sets of disentangled features 
% and letting $\mathbf{z}^{p}$ the invariant part of two augmentation views, we can extract the robust features and
and later utilizing them to reconstruct the two, we alleviate the feature suppression issue~\citep{CL_shortcut} at no cost of information sufficiency.  
We further add extra regularization to guarantee $\mathbf{z}^{p}$ does not collapse into shallow or partial features during the reconstruction process. We will introduce more details of GraphCV in the later contents.

\vspace{-0.1in}
\subsection{Disentanglement by Cross-View Reconstruction}
\vspace{-0.1in}
\label{sec: cross-view}

% In GCL, we usually leverage a graph encoder to 
% aggregate the feature of graph data as its representation. There are multiple choices of graph encoders in GCL, including GCN \citep{GCN} and GIN \citep{GIN}, etc.
In GCL, we usually leverage a graph encoder, such as a GCN \citep{GCN} or a GIN \citep{GIN}, to encode the graph data into its representation.
There are multiple choices of graph encoders in GCL, including GCN \citep{GCN} and GIN \citep{GIN}, etc. In this work, we adopt GIN as the backbone network $f$ for simplicity. Note that any other commonly-used graph encoders can also be adapted to our model. Given two augmentation views $t_{1}(G)$ and $t_{1}(G)$ (where $t_{1}(\cdot)$ and $t_{2}(\cdot)$ are IID sampled from the same family of augmentation $\mathcal{T}$), we firstly use the encoder $f(\cdot)$ to map them into a lower dimension hidden space for the two embeddings, $\mathbf{z}_{1}$ and $\mathbf{z}_{2}$. Instead of directly maximizing the agreement between the two representations $\mathbf{z}_{1}$ and $\mathbf{z}_{2}$, we further feed each of them into a pair decoders $(g_{p}$, $g_{c})$ (both of them are MLP-based networks or GNN) and optimize the two decoders to map each of the presentation into the two disentangled feature sub-spaces: 
\begin{equation}
\left[\mathbf{z}^{p} = g_{p}(f(t(G))) \text{,} \; \;
\mathbf{z}^{c} = g_{c}(f(t(G))) \right],
\end{equation}
where a pair of embeddings for both $t_{1}(G)$ and $t_{2}(G)$ are generated.
% through the procedure above. 
% Ideally, the mutual redundancy assumption between $(\mathbf{z}_{1}^{p}, \mathbf{z}_{2}^{p})$ can thus be guaranteed because $t_{1}(G)$ and $t_{1}(G)$ are augmented from the same original graph, and they naturally share the same predictive factors under the assumption of mutual redundancy stated in \ref{sec: CL}.
Ideally, $\mathbf{z}_{1}^{p}$ and $\mathbf{z}_{2}^{p}$ suffice the mutual redundancy assumption stated in \ref{sec: CL} because $t_{1}(G)$ and $t_{1}(G)$ are augmented from the same original graph, and thus naturally share the same predictive factors.

Here, we clarify the lower bound of the mutual information between one augmented view and the two mapped representations learned from the other augmented view in Theorem \ref{theorem: objective}.

\begin{theorem}
\label{theorem: objective}
Suppose $f(\cdot)$ is a GNN encoder as powerful as 1-WL test. Let $\mathbf{z}_{1}^{p}$ and $\mathbf{z}_{2}^{p}$ be specific to the predictive information of $G$, meanwhile $\mathbf{z}_{1}^{c}$ and $\mathbf{z}_{2}^{c}$ account for the non-predictive factors of $t_{1}(G)$ and $t_{2}(G)$. Then we have:
\begin{equation}
I\left(t_{1}(G) ; \mathbf{z}_{2}^{p}, \mathbf{z}_{2}^{c}\right) \geq I\left( \mathbf{z}_{1}^{p} ; \mathbf{z}_{2}^{p}\right) \text { where } G \in \mathcal{G} \text{ and } t_{1}(\cdot), t_{2}(\cdot) \in \mathcal{T}.
\nonumber
\end{equation}
\end{theorem}
The detailed proof is provided in Appendix \ref{appendix: proof}. 
Given the lower bound, we substitute the objective by the mutual information between the two representations in the predictive view ($\mathbf{z}_{1}^{p}$ and $\mathbf{z}_{2}^{p}$) to maximize the consistency between the information of the two views. Therefore, we derive the objective function ensuring view invariance as follows:
% Therefore, we can maximize the consistency between the representations of the two views by maximizing the mutual information of between $\mathbf{z}_{1}^{p}$ and $\mathbf{z}_{2}^{p}$. Therefore, we can derive our objective to ensure the view invariance as follow: 

\begin{equation}
\mathcal{L}_{\text{pre}}= \frac{1}{N} \sum_{i=1}^{N} \mathcal{L}_{\text{CL}}(\mathbf{z}_{1, i}^{p}, \mathbf{z}_{2, i}^{p}) ,
\label{eq: inv}
\end{equation}

% where $\mathcal{L}_{\text{CL}}(\cdot)$ denotes the contrastive objective and we adopt InfoNCE loss in this work \citep{InfoNCE}. 
where $\mathcal{L}_{\text{CL}}(\cdot)$ is the adopted InfoNCE loss~\citep{InfoNCE}.
To further pursue the feature disentanglement as illustrated in Figure~\ref{fig: mutual_info}(c), we thus propose the cross-view reconstruction mechanism. 
% To be specific, we utilize the representation pair $(\mathbf{z}^{p}, \mathbf{z}^{c})$ within and cross the augmentation views to recover the original representation $z$ so that the two objectives can be guaranteed simultaneously. Due to the reason that graph data is a kind of non-Euclidean structured data which cannot be represented in the euclidean space like the raw data in computer vision domain, we turn to infer the output of $\mathbf{z}=f(t(G))$ based on $(\mathbf{z}^{c}, \mathbf{z}^{p})$. 
To be specific, we would like the representation pair $(\mathbf{z}^{p}, \mathbf{z}^{c})$ within and cross the augmentation views be able to recover the raw data so that the two objectives can be approached simultaneously.
Due to the fact that graphs are non-Euclidean structured data, we instead try to recover $\mathbf{z}=f(t(G))$ given $(\mathbf{z}^{c}$ and $\mathbf{z}^{p})$. 

More specifically, we first perform the reconstruction within the augmentation view, namely mapping $(\mathbf{z}_{w}^{p}, \mathbf{z}_{w}^{c})$ to $\mathbf{z}_{w}$, where $w \in \{1, 2\}$ representing the augmentation view.  
% The optimal result of the this step is shown in Figure \ref{fig: mutual_info}(a), where the joint of $\mathbf{z}_{w}^{p}$ and $\mathbf{z}_{w}^{c}$ can cover all the information in its corresponding augmentation graph view $w$. 
% Since only $\mathbf{z}^{p}$ is involved in the contrastive loss in Equation \ref{eq: inv}, the graph encoder $f$ is thus not optimized to focus only on the easy-learned salient features and less powerful than 1-WL test. 
Then, we define the $(\mathbf{z}_{w^{\prime}}^{p}, \mathbf{z}_{w}^{c})$ as a cross-view representation pair and the reconstruction procedure is repeated on it to predict $\mathbf{z}_{w}$, aiming to ensure $\mathbf{z}^{p}$ and $\mathbf{z}^{c}$ is optimized to approximate mutual disentanglement, where $w=1, w^{\prime}=2$ or $w=2, w^{\prime}=1$. 
Intuitively, the reconstruction process is capable of separating the information of the shared features sets from the one resided in the unique feature sets between the two augmentation views. 
Since the two IID sampled augmentation operators ($t_{1}(\cdot)$ and $t_{2}(\cdot)$) are expected to preserve the predictive/rational features while varying the augmentation-related ones, we disentangle the rational features from $G$ according to the rationale discover studies~\citep{IRD} to ensure the features' robustness for downstream tasks.
Here, we formulate the reconstruction procedures as: 
\begin{equation}
\mathbf{z}_{w}^{r} = g_{r}\left( \mathbf{z}_{w}^{p} \odot \mathbf{z}_{w}^{c} \right) \text{,} \; \;
\mathbf{z}_{w}^{cr} = g_{r}\left( \mathbf{z}_{w^{\prime}}^{p} \odot \mathbf{z}_{w}^{c} \right) ,
\end{equation}
where $g_{r}$ is the parameterized reconstruction model and $\odot$ is the 
% pre-defined 
free-to-choose fusion operator, such as element-wise product or concatenation. The reconstruction procedures are optimized by minimizing the entropy $H\left( \mathbf{z}_{w} \mid \mathbf{z}_{w^{\prime}}^{p}, \mathbf{z}_{w}^{c} \right)$, where $w=w^{\prime}$ or $w \neq w^{\prime}$. 
Ideally, we reach the optimal sufficiency and disentanglement conditions illustrated in Figure \ref{fig: mutual_info} (b) and (c) iff $H\left(\mathbf{z}_{w} \mid \mathbf{z}_{w^{\prime}}^{p}, \mathbf{z}_{w}^{c}\right)=-\mathbb{E}_{p\left(  \mathbf{z}_{w}, \mathbf{z}_{w^{\prime}}^{p}, \mathbf{z}_{w}^{c} \right)}\left[\log p\left(\mathbf{z}_{w} \mid \mathbf{z}_{w^{\prime}}^{p}, \mathbf{z}_{w}^{c}\right)\right]=0$, 
where $\mathbf{z}_{w}$ is exactly recovered given its complementary representation and the predictive representation of any view. Nevertheless, the condition probability $p\left(\mathbf{z}_{w} \mid \mathbf{z}_{w^{\prime}}^{p}, \mathbf{z}_{w}^{c}\right)$ is 
% unknown for us, 
intractable, we hence use the variational distribution approximated by $g_{r}$ instead, denoted as $q\left(\mathbf{z}_{w} \mid \mathbf{z}_{w^{\prime}}^{p}, \mathbf{z}_{w}^{c}\right)$.
We provide the upper bound of $H\left( \mathbf{z}_{w} \mid \mathbf{z}_{w^{\prime}}^{p}, \mathbf{z}_{w}^{c} \right)$ in Theorem \ref{theorem: disentangle}.

\begin{theorem}
\label{theorem: disentangle}
Assume $q$ is a Gaussian distribution, $g_{r}$ is the parameterized reconstruction model which infers $\mathbf{z}_{w}$ from $\left( \mathbf{z}_{w^{\prime}}^{p}, \mathbf{z}_{w}^{c} \right)$. Then we have: 
\begin{equation}
H\left( \mathbf{z}_{w} \mid \mathbf{z}_{w^{\prime}}^{p}, \mathbf{z}_{w}^{c} \right) \leq \left\|\mathbf{z}_{w}-  g_{r}\left(\mathbf{z}_{w^{\prime}}^{p} \odot \mathbf{z}_{w}^{c}\right) \right\|_{2}^{2} \text { where } w=w^{\prime} \text{ or } w \neq w^{\prime}.
\nonumber
\end{equation}
\end{theorem}
The detailed proof is demonstrated in Appendix \ref{appendix: proof}.
Since we adopt two augmentation views, the objective function constraining representation disentanglement can be formulated as: 
\begin{equation}
\mathcal{L}_{\text{recon}} = \frac{1}{2N} \sum_{i=1}^{N} \sum_{w=1}^{2} \left[\left\|\mathbf{z}_{w, i}-  \mathbf{z}_{w, i}^{r} \right\|_{2}^{2}+\left\|\mathbf{z}_{w, i}- \mathbf{z}_{w, i}^{cr} \right\|_{2}^{2}\right].
\label{eq: disentanglement}
\end{equation}

\vspace{-0.15in}
\subsection{Adversarial Contrastive View}
\vspace{-0.1in}
\label{sec: adv-view}
%% TODO: 解释essential factors above
With the cross-view reconstruction mechanism above, the two learned representations stated above are optimized towards the disentangled manner. However, it is still necessary to further prevent the learned predictive representation from focusing on the partial features, because we do not have access to the explicit domain knowledge and such small scope will increase the risk of shortcut solution. 
% However, we still need to ensure $\mathbf{z}^{p}$ can maintain the global semantics of the graph instead of focusing on the partial salient features since we do not have access to the explicit domain knowledge. 
Therefore, we extend the Equation \ref{eq: inv} to three contrastive views and add an extra global view without topological perturbation as the third views to guarantee the learned $\mathbf{z}^{p}$ maintain the global semantics instead of partial or even trivial features, i.e., $\mathbf{z}_{1}^{p} \sim G$ and $\mathbf{z}_{2}^{p} \sim G$. 
During the experiments, we find an adversarial graph sample perturbed from original graph view can help the model achieve stronger robustness.
% A possible explanation is that there is still redundant information left in the shared part of the two augmentation views, especially when the implemented augmentations are moderate. Thus, an adversarial view can further eliminate the trivial information from the learned $\mathbf{z}^{p}$ for better generalization ability. 
A possible explanation is that there is still redundant information that is not predictive left in the shared information of the two $\mathbf{z}^{p}$'s in the two augmentation views, especially when the implemented augmentations are moderate. An adversarial view may further alleviate redundancy.
We define the adversarial objective as follows: 
\begin{equation}
\delta^{*} =
\underset{\left\|\delta \right\|_{\infty} \leq \epsilon}{\operatorname{argmax}}
\mathcal{L}_{\text{adv}}\left(t_{1}(G), t_{2}(G), G+\delta \right) ,
\label{eq: exp_adversarial}
\end{equation}
where the adversarial sample $G+\delta$ together with the two augmentation views, i.e., $t_{1}(G)$ and $t_{2}(G)$ are employed as the positive pair. Our crafted perturbation is spurred by recent work \citep{GASSL} that add perturbation $\delta$ on the output of first hidden layer $\mathbf{h}^{(1)}$, since it is empirically proved to generate more challenging views than adding perturbation on the initial node feature. Therefore, the adversarial contrastive objective is defined as:
\begin{equation}
\mathcal{L}_{\text{adv}} = \frac{1}{N} \sum_{i=1}^{N}
\underset{\delta^{*}}{\max}
\left[\mathcal{L}_{\text{CL}}\left(\mathbf{z}_{1, i}^{p}, G+\delta^{*} \right) + \mathcal{L}_{\text{CL}}\left(\mathbf{z}_{2, i}^{p}, G+\delta^{*} \right)\right].
\label{eq: adversarial}
\end{equation}
where the optimized perturbation $\delta^{\prime}$ is solved by projected gradient descent (PGD) \citep{PGD}.
Finally, we derive the joint objective of GraphCV by combining all of objectives above together. The joint objective is as follow:
% Given the graph $G \in \mathbf{G}$, the graph encoder $f$ and all the decoders $g$ can be optimized with the objective below: 
\begin{equation}
\min_{f, g} \mathbb{E}_{G \in \mathbf{G}} \left[\mathcal{L}_{\text{pre}} +\lambda_{r} \mathcal{L}_{\text{recon}} +\lambda_{a}
\underset{\left\|\delta \right\|_{\infty} \leq \epsilon}{\max}
\mathcal{L}_{\text{adv}} \right] \\,
\end{equation}
where $\lambda_{r}$ and $\lambda_{a}$ are the coefficients to balance the magnitude of each loss term. 
% With the combined objective and appropriate magnitude of each loss item, 
Our proposed model is able to learn optimal representation illustrated in Figure \ref{fig: mutual_info}(c) with the joint objective.

\vspace{-0.15in}
\section{Experiments}
\vspace{-0.1in}
In this section, we demonstrate the empirical evaluation results of GraphCV on public graph benchmark datasets under different settings. 
% under both unsupervised and semi-supervised settings. 
Ablation study and hyper-parameter analysis are also conducted to evaluate the effectiveness of the designs in GraphCV. We further compare the robustness of GraphCV with the adversarial training-based GCL method. 
More content about dataset statistics, training details and other empirical analysis are provided in the Appendix. 

\vspace{-0.1in}
\subsection{Experimental Setups}
\vspace{-0.1in}

\noindent
\textbf{Datasets. }For unsupervised learning setting, we evaluate our model on five graph benchmark datasets from the field of bioinformatics, including MUTAG, PTC-MR, NCI1, DD, and PROTEINS, and other four from the field of social network, which are COLLAB, IMDB-B, RDT-B, and IMDB-M, for the task of graph-level property classification. 
For the transfer learning setting, we follow previous work~\citep{GraphCL, GraphLoG} to pretrain our model on the ZINC-2M dataset, which cotains 2 million unlabeled molecule graphs sampled from MoleculeNet~\citep{moleculenet}, then evaluate its performance on eight binary classification datasets from chemistry domain, where the eight datasets are splitted according to the scaffold to simulate the out-of-distribution scenario in real-world. 
Additionally, We use ogbg-molhiv from Open Graph Benchmark Dataset \citep{OGB} to evaluate our model over large-scale dataset under semi-supervised setting. More details about dataset statistics are included in Appendix \ref{appendix: dataste_statistics}.

\noindent
\textbf{Baselines. }Under the unsupervised representation learning setting, we compare GraphCV with the eight SOTA self-supervised learning methods GraphCL \citep{GraphCL}, InfoGraph\citep{InfoGraph}, MVGRL \citep{ContrastMultiView},  AD-GCL\citep{AD-GCL}, GASSL\citep{GASSL}, InfoGCL\citep{InfoGCL}, RGCL~\citep{RGCL} and DGCL\citep{DGCL}, as well as three classical unsupervised representation learning methods, including node2vec \citep{node2vec}, 
% sub2vec \citep{sub2vec}, 
graph2vec \citep{graph2vec}, and GVAE\citep{VGAE}. Besides, we employ AttrMasking~\citep{pretrain_strategies}, ContextPred~\citep{pretrain_strategies}, GraphCL~\citep{GraphCL}, GraphLoG~\citep{GraphLoG}, AD-GCL~\citep{AD-GCL} and RGCL~\citep{RGCL} as baselines to evaluate the effectiveness of our proposed GraphCV under transfer learning setting.  

\noindent
\textbf{Evaluation Protocol. }For unsupervised setting,  we follow the evaluation protocols in the previous works \citep{InfoGraph, GraphCL, DGCL} to verify the effectiveness of our model. 
% The learned representation is fine-tuned by a linear SVM classifier for task-specific prediction.
The mean test accuracy score evaluated by a 10-fold cross validation with standard deviation of five random seeds as the final performance. 
For transfer learning setting, we follow the finetuning procedures of previous work~\citep{GraphCL, GraphLoG} and report the mean ROC-AUC scores with standard deviation of 10 repeated runs on each downstream datasets. 
In addition, we follow the setting of semi-supervised representation learning from GraphCL on the ogbg-molhiv dataset, with the finetune label rates as 1\%, 10\%, and 20\%. The final performance is reported as the mean ROC-AUC of five initialization random seeds

% \textbf{Implementation Details. }We implement our framework with PyTorch and PyGCL library \citep{PyGCL}. We choose GIN \citep{GIN} as the backbone graph encoder and the model is optimized through Adam optimizer. There are two specific hyperparameters in our model, namely $\lambda_{r}$ and $\lambda_{a}$, the search space of them are 0.0 to 10.0 and 0.0 to 1.0, respectively. More details about implementation details is provided in Appendix~\ref{appendix: implement}. 
\label{sec: exp_setup}

\begingroup
\begin{table*}[t]
\centering
\vspace{-0.4in}
\caption{Overall comparison on multiple graph classification benchmarks under unsupervised learning setting. Results are reported as mean±std\%,  the best performance is bolded and runner-ups are underlined. "-" indicates the result is not reported in original papers.}
% \vspace{-0.15in}
\label{tab:res}
\setlength{\tabcolsep}{3pt}
\begin{adjustbox}{width=\textwidth,center}
% \begin{tabular}{p{0.2\textwidth}cp{0.2\textwidth}ccccccccc}
\begin{tabular}{ccccccccccc}
\toprule
  & \textbf{MUTAG} & \textbf{PTC-MR} & \textbf{COLLAB} & \textbf{NCI1} & \textbf{PROTEINS}  & \textbf{IMDB-B}   & \textbf{RDT-B}  & \textbf{IMDB-M} & \textbf{DD}\\
\midrule
node2vec&	72.6±10.2& 58.6±8.0  &   -&	54.9±1.6 &	57.5±3.6&	-&	-&	-&	-\\
% sub2vec&	61.1±15.8&	60.0±6.4&	-&	52.8±1.5& 53.0±5.6 & 	55.3±1.5  & 71.5±0.4  &		36.7±0.8 & -\\
graph2vec&	83.2±9.3&	60.2±6.9 &	-&	73.2±1.8& 73.3±2.1 & 	71.1±0.5  & 75.8±1.0   &	50.4±0.9  & -\\
InfoGraph&	89.0±1.1&	61.7±1.4  &	70.7±1.1&	76.2±1.1 & 74.4±0.3 & 	73.0±0.9  & 82.5±1.4   &	49.7±0.5  & 72.9±1.8 \\
VGAE&	87.7±0.7&	61.2±1.8   & -&	-& - & 	70.7±0.7  & 87.1±0.1   &	49.3±0.4  & - \\
MVGRL&	89.7±1.1&	62.5±1.7   & -&	-& - & 	74.2±0.7   & 84.5±0.6    &	51.2±0.5  & - \\
\midrule
GraphCL & 86.8±1.3 & 63.6±1.8 &	71.4±1.2 &  77.9±0.4 &	74.4±0.5 &  71.1±0.4 &	89.5±0.8 &	- &  \underline{78.6±0.4}\\
InfoGCL&	91.2±1.3& 63.5±1.5 &   80.0±1.3&	80.2±0.6&	-&	75.1±0.9&	-&	51.4±0.8& -\\
DGCL&	\underline{92.1±0.8}&	\underline{65.8±1.5}&	\textbf{81.2±0.3}&	\underline{81.9±0.2}&	\underline{76.4±0.5}&	\textbf{75.9±0.7}&	\underline{91.8±0.2}&	\underline{51.9±0.4}& -	\\
AD-GCL&	89.7±1.0& - &   73.3±0.6&	69.7±0.5&	73.8±0.5&	72.3±0.6&	85.5±0.8&	49.9±0.7&	75.1±0.4\\
RGCL&	87.7±1.0& - &   70.9±0.7&	78.1±1.1&	75.0±0.4&	71.9±0.8&	90.3±0.6&	-&	78.9±0.5\\
GASSL&	90.9±7.9&	64.6±6.1&	78.0±2.0&	80.2±1.9& - & 	74.2±0.5 & - &		51.7±2.5 & -\\

\midrule
\textbf{GraphCV}&	\textbf{92.3±0.7}&	\textbf{67.4±1.3}&	\underline{80.5±0.5}&	\textbf{82.0±1.0}&	\textbf{76.8±0.4}&	\underline{75.6±0.4}&	\textbf{92.4±0.9}&	\textbf{52.2±0.5}&	 \textbf{80.5±0.5}\\
\bottomrule
\end{tabular}
\end{adjustbox}
\vspace{-0.2in}
\end{table*}
\endgroup

\vspace{-0.1 in}
\subsection{Overall Performance Comparison}
\vspace{-0.1 in}
\label{sec: exp_res}
\textbf{Unsupervised learning. }The overall performance comparison is shown in Table~\ref{tab:res} and we can have three observations: (1) The GCL-based methods generally yield higher performances than classical unsupervised learning methods, indicating the effectiveness of utilizing instance-level supervision; (2) RGCL, AD-GCL, and GASSL achieve better performances than GraphCL, which empirically proves the conclusion that InfoMax object could bring overwhelmed redundant information and thus suffer from feature suppression issue; (3) Our proposed GraphCV and DGCL consistently outperform other baselines, proving the advantage of disentangled representation. More importantly, GraphCV achieves state-of-the-art results on most of the datasets, demonstrating the model effectiveness. 
% to learn minimal yet sufficient representations.

%% TODO: check result
% \vspace{-0.05 in}
\textbf{Transfer learning. } Table \ref{tab: transfer} demonstrates the experimental results under transfer learning setting, where No Pre-Train skips self-supervised pre-training process on the ZINC-2M dataset for model initialization before finetune.
% Except for the baselines mentioned above, we also include No Pre-Trained GIN as the baseline (i.e., without self-supervised learning task on the ZINC-2M dataset for model initialization before finetune). 
It is noteworthy that some strong baselines (AttrMasking and ContextPred) are trained under the guidance of domain knowledge. Despite in lacking of such domain knowledge, our model still outperforms all the other baselines on 3 out 8 datasets and achieve highest average performance. More importantly, JOAO, RGCL and our proposed GraphCV are all developed from GraphCL, but achieve higher average performance than GraphCL. This observation further empirically prove the poisoning effect of biased information and the necessity to to suppress them.
% necessity to alleviate the feature suppression issue and the superiority to reduce redundant information in learned representation. 

\begingroup
\begin{table*}[t]
\centering
\vspace{-0.0in}
\caption{Overall comparison on multiple graph classification benchmarks under transfer learning setting. Results are reported as mean±std\%,  the best performance is bolded and runner-ups are underlined.}
% \vspace{-0.15in}
\label{tab: transfer}
\setlength{\tabcolsep}{3pt}
\begin{adjustbox}{width=\textwidth,center}
\begin{tabular}{ccccccccccc} 
\toprule
  & \textbf{BBBP} & \textbf{Tox21} & \textbf{ToxCast} & \textbf{SIDER} & \textbf{ClinTox}  & \textbf{MUV}   & \textbf{HIV}  & \textbf{BACE} & \textbf{Avg} \\
\midrule
No Pre-Train &65.8±4.5    & 74.0±0.8 &	63.4 ±0.6&	57.3±1.6&	58.0±4.4&	71.8±2.5&	75.3±1.9&	70.1±5.4 &	67.0\\
AttrMasking &64.3±2.8    & \textbf{76.7±0.4} &	\textbf{64.2±0.5}&	61.0±0.7&	71.8±4.1&	74.7±1.4&	77.2±1.1&	79.3±1.6 &	71.1\\
ContextPred &68.0±2.0	&75.7±0.7	&63.9±0.6	&60.9±0.6	&65.9±3.8	&75.8±1.7	&77.3±1.0	&79.6±1.2 &	70.9\\
GraphCL &69.5±0.5   & 75.4±0.9 &	63.8±0.4&	60.8±0.7&	70.1±1.9&	74.5±1.3&	77.6±0.9&	78.2±1.2 &	70.8\\
GraphLoG & \textbf{72.5±0.8}    & 75.7±0.5 &	63.5±0.7&	61.2±1.1&	76.7±3.3&	76.0±1.1&	77.8±0.8&	\textbf{83.5±1.2} &	73.4\\
JOAO &70.2±1.0    & 75.0±0.3 &	62.9±0.5&	60.0±0.8&	81.3±2.5&	71.7±1.4&	76.7±1.2&	51.5±0.4 &	71.9\\
RGCL &71.4±0.7    & 75.2±0.3 &	63.3±0.2&	\underline{61.4±0.6}&	\underline{83.4 ±0.9}&	\textbf{76.7 ±1.0}&	\underline{77.9±0.8}&	76.03±0.8 &	\underline{73.2}\\
\midrule
\textbf{GraphCV}&	\underline{71.6±0.6}&	\underline{75.7±0.6}&	\underline{63.2±0.5}&	\textbf{62.2±0.7}&	\textbf{83.6±1.5}&	\underline{76.4±0.8}&	\textbf{77.9±1.0}&	\underline{80.8±1.8} &	\textbf{73.9}\\
\bottomrule
\end{tabular}
\end{adjustbox}
\vspace{-0.1in}
\end{table*}
\endgroup

\begin{wrapfigure}{r}{6.5cm}
    \centering
    \vspace{-0.25in}
    \includegraphics[width=1\linewidth]{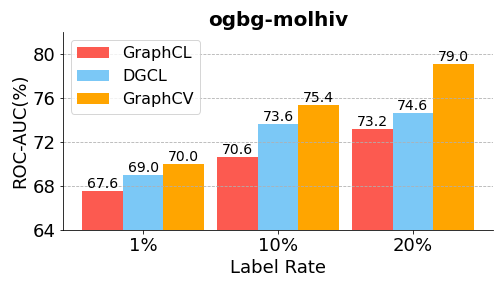}
    \vspace{-0.4in}
    \caption{Performance comparison of semi-supervised learning on ogbg-molhiv.}
        \vspace{-0.05in}
    \label{fig: molhiv}
\end{wrapfigure}

% \vspace{-0.1 in}
\noindent
\textbf{Semi-supervised learning. }The experimental results are shown in Figure~\ref{fig: molhiv}. It is obvious that our model gains significant improvements under the three label-rate fine-tuning settings. We also notice that as the label rate increases, the amount of improvement increases as well (1\%, 1.8\%, and 4.4\% for label rate 1\%, 10\%, and 20\%, respectively). A possible explanation could be that more trainable data could bring more redundant information, thereby further deteriorate the feature suppression issue.
Therefore, removing redundant information causes a higher performance boost.

\vspace{-0.1 in}
\subsection{Ablation Study}
\vspace{-0.1 in}

To further verify the effectiveness of different modules in GraphCV, we perform ablation studies on each one of the module by creating two model variants: (1) \textbf{w/o CV Recon}, the cross-view reconstruction process is discarded; (2) \textbf{w/o Adv. Training}, the third adversarial view is discarded.
The comparison results are shown in Table~\ref{tab:abl}.
% \vspace{-0.1 in}
% \begin{itemize}[leftmargin=*]
%     \item \textbf{w/o Intra-view Recon. }Reconstruction is only executed within the cross view i.e., $w \neq w^{\prime}$.
%     \item \textbf{w/o Inter-view Recon. }Reconstruction is only executed within the same view i.e., $w = w^{\prime}$.
%     % \item \textbf{w/o CV Recon. } The cross-view reconstruction process is discarded.
%     \item \textbf{w/o Adv. Training. }Adversarial view is discarded in the contrastive loss.
% \end{itemize}
% \vspace{-0.1 in}
We can observe from Table~\ref{tab:abl} that our model with the combination of cross-view reconstruction and adversarial training module outperforms all of the variants. Omitting the reconstruction process could cause the failure to optimize the representation in a disentangled manner illustrated in Figure \ref{fig: mutual_info}(c), thereby the learned representation still suffer from features suppression issue. 
% We can not guarantee the representation disentanglement assumption if we skip the inter-view reconstruction, and the sufficiency assumption may not hold if we abandon intra-view reconstruction. Either way, the augmentation-invariant representations may suffer from enormous information loss during the contrastive learning and further lead to the performance deterioration. 
Compared with our model, the variant w/o Adv. Training may lead to representation collapse and bring extra redundant information, therefore resulting in sub-optimal performances in downstream tasks. 
% The relatively larger performance deterioration for  the two variants w/o Intra-view Recon and w/o Inter-view suggests the rule "better than nothing". That is, having redundant information is better than having it partially.

\begingroup
\begin{table*}[h]
\centering
\vspace{-0.1in}
\caption{Overall comparison of the model variants’ performance. Results are reported as mean±std\%, the best performance is bolded.}
% \vspace{-0.15in}
\label{tab:abl}
\setlength{\tabcolsep}{3pt}
\begin{adjustbox}{width=\textwidth,center}
\begin{tabular}{ccccccccccc} 
\toprule
  & \textbf{MUTAG} & \textbf{PTC-MR} & \textbf{COLLAB} & \textbf{NCI1} & \textbf{PROTEINS}  & \textbf{IMDB-B}   & \textbf{RDT-B}  & \textbf{IMDB-M} & \textbf{DD}\\
\midrule
% w/o Intra Recon &91.5±1.2    & 65.8±1.3 &	78.4±0.7&	79.6±0.7&	75.6±0.5&	74.2±0.8&	92.0±0.4&	51.5±0.4&	55.8±0.6&	79.3±0.7\\
% w/o Inter Recon &91.0±0.9    & 64.7±1.4 &	78.0±0.8&	78.7±1.2&	74.9±0.7&	75.0±0.6&	91.1±0.7&	50.8±0.2&	55.6±0.4&	79.0±0.8\\
w/o CV Recon &91.0±0.9    & 64.7±1.4 &	78.0±0.8&	78.7±1.2&	74.9±0.7&	75.0±0.6&	91.1±0.7&	51.7±0.6&	79.0±0.8\\
w/o Adv. Training &92.1±0.6	&66.8±0.5	&76.5±0.6	&81.2±0.9	&76.0±0.3	&75.1±0.6	&92.2±1.0	&50.8±0.4	&80.1±0.6\\
\textbf{GraphCV}&	\textbf{92.3±0.7}&	\textbf{67.4±0.5}&	\textbf{80.5±0.5}&	\textbf{82.0±1.0}&	\textbf{76.8±0.4}&	\textbf{75.6±0.4}&	\textbf{92.5±0.9}&	\textbf{52.2±0.5}&	\textbf{80.5±0.5}\\
\bottomrule
\end{tabular}
\end{adjustbox}
\end{table*}
\endgroup

\vspace{-0.15in}
\subsection{Robustness and Hyper-parameter Analysis}
% \vspace{-0.1 in}

%  \vspace{-0.1in}
\begin{figure*}[h]
  \centering
  \includegraphics[width=\linewidth]{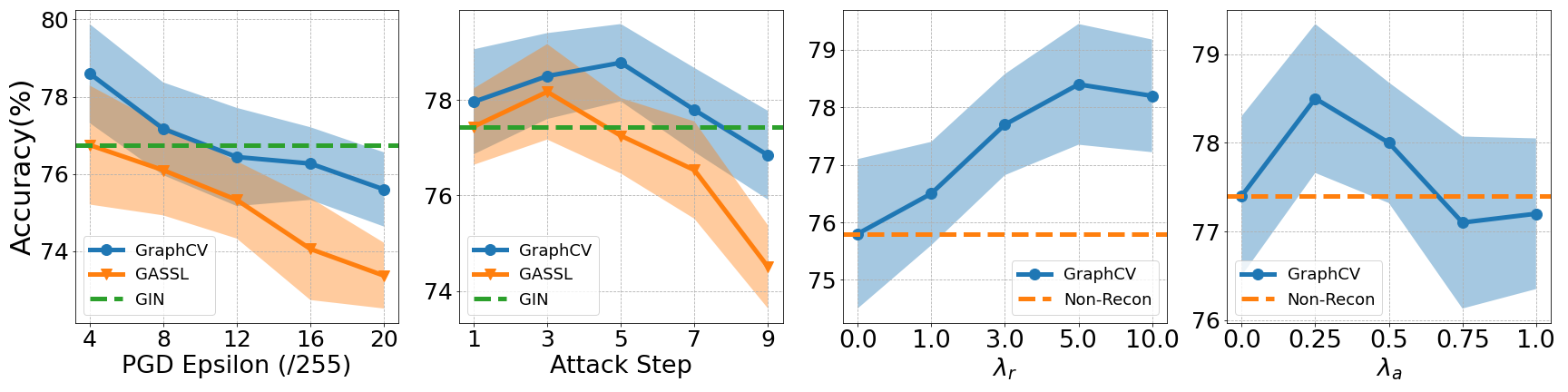}
 \vspace{-0.25in}
  \caption{The model performance under different perturbation bound, attack step and analysis the sensitivity of the two important hyper-parameters (i.e., $\lambda_{r}$ and $\lambda_{a}$).}
  \vspace{-0.15in}
  \label{fig: aug}
\end{figure*}

In this section, we firstly conduct extra experiments on ogbg-molhiv dataset to evaluate the representation robustness under aggressive augmentation and perturbation. 
The results are shown in left two subplots of Figure~\ref{fig: aug}, we compare our method with GASSL under different perturbation bounds and attack steps to demonstrate its robustness against adversarial attacks. 
Since both our model and GASSL use GIN as the backbone network, we hereby add the performance of GIN as the compared baseline.
Although aggressive adversarial attacks can largely deteriorate the performance, our proposed GraphCV still achieves more robust performance than GASSL. 
In the right two subplots, we analysis the model sensitivity of the two important hyper-parameters in our model, $\lambda_{r}$ and $\lambda_{a}$. The consistent superiority of different values over the initial point (i.e., $\lambda_{r}$, $\lambda_{a}=0$) prove the effectiveness of our design once again. We can also observe that the appropriate range of the two hyper-parameters are 5.0 to 10.0 and 0.0 to 0.5, respectively. Depend on the datasets size and attributes, the range can have some variance, we suggest to finetune the two hyper-parameters around 10.0 and 0.25 to find the appropriate values when adopting our model to a new datasets. More experiments and discussion about hyper-parameter sensitivity in provided in the Appendix \ref{sec: appendix_hyper}. Besides, we also conduct extra experiments to analyze the disentanglement of $\mathbf{z}^{p}$ and $\mathbf{z}^{c}$ in Appendix \ref{appendix: disentangle}.

% verses edge perturbation and attribute masking strengths, respectively. Specifically, we keep the GraphCL and our proposed GraphCV under the same hyperparameter setting and set the $\lambda_{r}$ and $\lambda_{a}$ of GraphCV as 10.0 and 0.5, respectively.
% From the results we can see that GraphCV not only consistently outperforms GraphCL but also is less affected by larger augmentation strengths. 

% 
% \vspace{-0.2 in}
\section{Related Work}
% \vspace{-0.1in}
\textbf{Graph contrastive learning. }Contrastive learning is firstly proposed in the compute vision field \citep{SimCLR} and raises a surge of interests in the area of self-supervised graph representation learning for the past few years. The principle behind contrastive learning is to utilize the instance-level identity as supervision and maximize the consistency between positive pairs in hidden space through designed contrast mode. Previous graph contrastive learning works generally rely on various graph augmentation (transformation) techniques \citep{DGI, GCC, MVGRL, GraphCL, InfoGraph} to generate positive pair from original data as similar samples. Recent works in this field try to improve the effectiveness of graph contrastive learning by finding more challenge view \citep{AD-GCL, InfoGCL, JOAO} or adding adversarial perturbation \citep{GASSL}. However, most of the existing methods contrast over entangled embeddings, where the complex intertwined information may pose obstacles to extracting useful information for downstream tasks. Our model is spared from the issue by contrasting over disentangled representations.

\noindent
\textbf{Disentangled representation learning on graphs. }Disentangled representation learning arises from the computer vision field \citep{hsieh_learning_2018, zhao_learning_2021} to disentangle the heterogeneous latent factors of the representations, and therefore making the representations more robust and interpretable \citep{RLReview}.
% This idea has been explored in both the computer vision and graph representation learning fields, supervised and unsupervised \citep{hsieh_learning_2018, zhao_learning_2021, InfoGAN, DisenGCN, DGCL}. Unlike \citep{hsieh_learning_2018} which specifically disentangles the view and pose factors, \citep{DGCL} implicitly disentangles the structured information via the assumption of channels. Our method however, specifically disentangles the augmentation related factors and those relative to the graph-level properties in an unsupervised way.
This idea has now been widely adopted in graph representation learning. \citep{IPGDN, DisenGCN} utilizes neighborhood routing mechanism to identify the latent factors in the node representations. Some other generative models \citep{VGAE, GraphVAE} utilize Variational Autoencoders to balance reconstruction and disentanglement. Recent work \citep{DGCL} outspreads the application of disentangled representations learning in self-supervised graph learning by contrasting the factorized representations. Although these methods gain significant benefit from the representation disentanglement, the underlined excessive information could still overload the model, thus resulting in limited capacities. Our model targets the issue by removing the redundant information that is considered irrelevant to the graph property. 

\noindent
\textbf{Graph information bottleneck. }The Information bottleneck (IB) \citep{IB} has been widely adopted as a critical principle of representation learning. A representation contains minimal yet sufficient information is considered to be in compliance with the IB priciple and many works \citep{VIB, blackbox, robust_rep} have empirically and theoretically proved that representation agree with IB principle is both informative and robust. Recently, IB principle is also borrowed to guide the representation learning of graph structure data. Current methods \citep{GIB, InfoGCL, AD-GCL, RGCL} usually propose different regularization designs to learn compressed yet informative representations in accordance with IB principle. We follow the information bottleneck to learn the expressive and robust representation from disentangled information in this work. 

\vspace{-0.05in}
\section{Conclusion}
\vspace{-0.1in}
\label{sec: conclusion}
 In this paper, we study the feature suppression problem in representation learning. To avoid the predictive features being suppressed in learned representation, we propose a novel model, namely GraphCV, which is designed in accordance with the information bottleneck principle. The cross-view reconstruction in GraphCV can disentangle those more robust and transferable features from those easily-disturbed ones.
%  so that the information entanglement brought by augmentations will not cause the loss of predictive information during contrastive learning. 
 Meanwhile, we also add an adversarial view as the third view of the contrastive learning to to guarantee the global semantics and further enhance representation robustness. In addition, we theoretically analyze the effectiveness of each component in our model and derive the objective based on the analysis. Extensive experiments on multiple graph benchmark datasets and different settings prove the ability of GraphCV to learn robust and transferable graph representation. In the future, we can explore how to come up with a practical objective to further decrease the upper bound of the mutual information between the disentangled representations and try to utilize more efficient training strategy to make the proposed model more time-saving on large-scale graphs. 

\newpage
\section*{Ethics Statement}
This idea is proposed to solve the general graph learning problem, so we believe there should exist no ethical concern applicable to our work. Any unethical application that benefits from our work is against our initial intent. 

\section*{Reproducibility Statement}
We provide the source code along with the submission in the supplementary materials for reproducibility. The source code and all the implementation details will be open to public once upon the acceptance of this paper. 

% \bibliography{iclr2023_conference}
% \bibliographystyle{iclr2023_conference}

\appendix

\section{Out-of-distribution Scenario on Graph}
\label{appendix: ood_case}

\begin{figure*}[h]
    \centering
    \vspace{-0.1in}
    \includegraphics[width=1\linewidth]{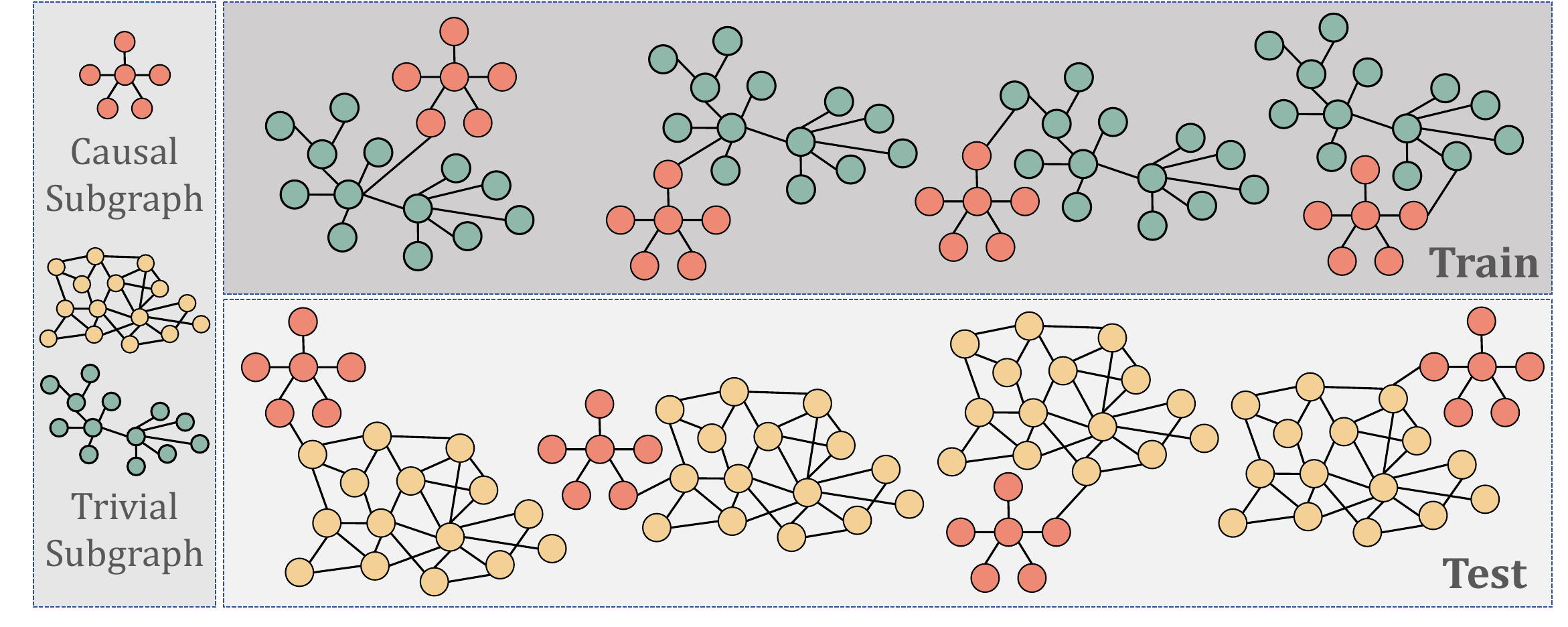}
    \vspace{-0.3in}
    \caption{An out-of-distribution situation in molecule graph prediction task. The casual functional sub-structure (red) are spuriously correlated with different trivial sub-structures in training and test set. The statistical correlations can lead to poor robustness and transferability.}
    \vspace{-0.1in}
    \label{fig: ood}
\end{figure*}

In this section, we will illustrate the out-of-distribution scenario in graph learning task. During molecule property study, A specific kind of property (e.g., toxicity and lipophilicity) of a molecule is usually dependent on if it has corresponding sub-structures (termed as functional group). For example, hydrophilic molecules usually have oxhydryl group ($-OH$)
Therefore, a well-trained GNN model on molecule graph prediction task is capable to reflect the sub-structure information in the graph representation. However, it is usually the case in real-world scenario that the predictive functional group is usually accompanied by some irrelevant groups in some environments, thus causing spurious correlations. This correlation usually lead to poor generalization performance when the model is evaluated on another environment with different spurious correlation. Figure \ref{fig: ood} intuitively demonstrates this kind of scenario, where the red subgraph is the feature we can rely on to make casual prediction. But it usually show up with green subgraph that do not serve as the functional graph of the property in training set. Consequently, the model are easily misguided that the green subgraph is an important indicator of the property. When we evaluate the model on the testing set where the casual graph is correlated with another kind of group (yellow subgraph), there usually exists a huge gap between its performances on the two sets. 

\section{Discussion on Feature Suppression}
\label{appendix: feature_suppression}

In this section, we will follow the previous works~\citep{CL_FS, CL_shortcut} to present a more formal definition of the feature suppression and clarify its relation with contrastive learning.

First of all, we assume graph data $G$ has $n$ feature sub-spaces, $G^1, \ldots, G^n$, where each $G^i \in G$ corresponding to a distinct feature of $G$. To quantify the relation between $G$ and its feature sub-spaces, we need to measure the conditional probability of $G$ given a specific kind of feature sub-space $G^i$ ($i \subseteq[n]$), denoted as $p(G \mid G^i)$. Finally, we define a an injective map $g: G^i \rightarrow G$ produce observation $G=g(G^i)$. Due to the reason that $G^i$ is not explicit, so we aim to train a encoder $f: G_{i} \rightarrow \mathbb{R}^{d}$ to map input graph data $G$ into a latent space to extract useful high-level information $\mathbf{z}^{i}$ corresponding to each feature sub-space $G^i$ of input data $G$ during cotrastive learning. Therefore, we use $p(G \mid \mathbf{z}^{i})$ as the approximated value of the measurement $p(G \mid G^{i})$. Then we have, 
\begin{itemize}
    \item For any feature sub-space $G^i$ and its complementary feature sub-subspace $G^{\bar{i}}$, $f$ suppress feature $i \subseteq[n]$ if we have $p(G \mid \mathbf{z}^{i}) = p(G \mid \mathbf{z}^{\bar{i}})$
    \item For any feature sub-space $G^i$ and its complementary feature sub-subspace $G^{\bar{i}}$, $f$ distinguish feature $i \subseteq[n]$ if $p(G \mid \mathbf{z}^{i})$ and $p(G \mid \mathbf{z}^{\bar{i}})$ have disjoint support. 
\end{itemize}
To sum up, a feature is suppressed if it does not make any difference to the instance discrimination. One of the common acknowledgements for unsuprevised learning strategy is that it can usually produce representation with uniform feature space distribution due to the lack of supervision, i.e., every feature sub-space is equally treated without feature suppression. However, it could not be the situation in contrastive learning. Taking the commonly used InfoNCE~]\citep{InfoNCE} as an example, it can be divided into two parts, i.e. align term and uniform term~\citep{SimCLR}, as follow:

\begin{equation}
\tau \mathcal{L}^{\mathrm{InfoNCE}}=\underbrace{-\frac{1}{m} \sum_{i, j} \operatorname{sim}\left(\boldsymbol{z}_i, \boldsymbol{z}_j\right)}_{\mathcal{L}_{\text {alignment }}}+\underbrace{\frac{\tau}{m} \sum_i \log \sum_{k=1}^{2 m} \mathbf{1}_{[k \neq i]} \exp \left(\operatorname{sim}\left(\boldsymbol{z}_i, \boldsymbol{z}_k\right) / \tau\right)}_{\mathcal{L}_{\text {uniform}}}
\end{equation}
Aligning the positive pair will distinguish the shared feature subspace $G^{i}$. Meanwhile, there also exits random negative samples that might own same factors in $G^{i}$, so the uniform term might suppress the feature sub-space $G^{i}$. 
Therefore, for any feature $i \subseteq[n]$, the optimization process can either suppress or distinguish it, but both of them can reach to lower contrastive loss. From the analysis we can derive the conclusion mentioned in Section \ref{sec: intro} that lower contrastive loss might not yield better performance.

\section{Summary of Datasets}
\label{appendix: dataste_statistics}
In this work, we use nine datasets from TU Benchmark Datasets \citep{TUDataset} to evaluate our proposed GraphCV under unsupervised setting, where five of them are biochemical datasets and the other four belong to social network datasets.
We also utilize the ogng-molhiv dataset from Open Graph Benchmark (OGB) \citep{OGB} to further evaluate GraphCV under semi-supervised setting.  Besides, the datasets sampled from MoleculeNet~\citep{moleculenet} are employed to evaluate our model under transfer learning setting. The statistics of these datasets are shown in Table \ref{tab: tu_data_stat} and \ref{tab: transfer_data_stat}. 

\begin{table*}[h]
\centering
\begin{tabular}{ccccccc} 
\toprule
  \textbf{Dataset} & \textbf{\#Graphs} & \textbf{Avg \#Nodes} & \textbf{Avg \#Edges} & \textbf{\#Class} & \textbf{Metric}  & \textbf{Category} \\
\midrule
MUTAG &188    & 17.93 &	19.79&	2&	Accuracy&	biochemical\\
PTC-MR &344  &14.29  &  14.69&  2&	 Accuracy&	biochemical\\
PROTEINS &1,113	&39.06	&72.82	&2	&Accuracy	&biochemical\\
NCI1 &4,110	&29.87	&32.30	&2	&Accuracy	&biochemical\\
DD &1,178	&284.32	&715.66	&2	&Accuracy	&biochemical\\
COLLAB &5,000	&74.49	&2457.78	&3	&Accuracy	&social network\\
IMDB-B &1,000	&19.77	&96.53	&2	&Accuracy	&social network\\
RDT-B &2,000	&429.63	&497.75	&2	&Accuracy	&social network\\
IMDB-M &1,500	&13.00	&65.94	&3	&Accuracy	&social network\\
ogbg-molhiv &41,127	&25.50 		&27.50	&2	&ROC-AUC	&MoleculeNet\\
\bottomrule
\end{tabular}
\caption{Statistics of TU-datasets and OGB dataset. }
\label{tab: tu_data_stat}
\end{table*}

\begin{table*}[h]
\centering
\begin{tabular}{ccccccc} 
\toprule
  \textbf{Dataset} & \textbf{\#Graphs} & \textbf{Avg \#Nodes} & \textbf{Avg Degree} & \textbf{\#Tasks} & \textbf{Metric}  & \textbf{Category} \\
\midrule
ZINC-2M &2,000,000    & 26.62 &	57.72&	-&	-&	biochemical\\
BBBP &2,039  &24.06  &  51.90&  1&	 ROC-AUC&	biochemical\\
Tox21 &7,813	&18.57	&38.58	&12	&ROC-AUC	&biochemical\\
ToxCast &8,576	&18.78	&38.62	&617	&ROC-AUC	&biochemical\\
SIDER &1,427	&33.64	&70.71	&27	&ROC-AUC	&biochemical\\
ClinTox &1,477	&26.15	&55.76	&2	&ROC-AUC	&biochemical\\
MUV &93,087	&24.23	&52.55	&17	&ROC-AUC	&biochemical\\
HIV &41,127	&25.51	&54.93	&1	&ROC-AUC	&biochemical\\
BACE &1,513	&34.08	&73.71	&1	&ROC-AUC	&biochemical\\
\bottomrule
\end{tabular}
\caption{Statistics of MoleculeNet datasets. }
\label{tab: transfer_data_stat}
\end{table*}

All of the eleven datasets are public available, we attach attach their links as follow: 
\begin{itemize}
    \item TU datasets: \url{https://chrsmrrs.github.io/datasets/docs/datasets/}
    \item MoleculeNet datasets: \url{http://snap.stanford.edu/gnn-pretrain/}
    \item ogbg-molhiv dataset: \url{https://ogb.stanford.edu/docs/graphprop/#ogbg-mol} 
\end{itemize}

\section{Implementation Details}
\label{appendix: implement}
All experiments are conducted with the following settings:
\begin{itemize}
    \item Operating System: Ubuntu 18.04.5 LTS
    \item CPU: AMD(R) Ryzen 9 3900x
    \item GPU: NVIDIA GeForce RTX 2080ti
    \item Software: Python 3.8.5; Pytorch 1.10.1; PyTorch Geometric 2.0.4; PyGCL 0.1.2; Numpy 1.20.1; scikit-learn 0.24.1. 
\end{itemize}
We implement our framework with PyTorch and PyGCL library \citep{PyGCL}. We choose GIN \citep{GIN} as the backbone graph encoder and the model is optimized through Adam optimizer. 
% There are two specific hyperparameters in our model, namely $\lambda_{r}$ and $\lambda_{a}$, the search space of them are 0.0 to 10.0 and 0.0 to 1.0, respectively. 
% More details about implementation details is provided in Appendix~\ref{appendix: implement}.
We choose GIN \citep{GIN} as the backbone graph encoder and the model is optimized through Adam optimizer. We follow \citep{GraphCL, GASSL, DGCL} to employ a linear SVM classifier for downstream task-specific classification. The graph augmentation operations used in our work are same as \citep{GraphCL}, including node dropping, edge perturbation, attribute masking and subgraph sampling, all of them are borrowed from the implementation of \citep{PyGCL}.  There are two specific hyper-parameters in our model, namely $\lambda_{r}$ and $\lambda_{a}$, the search space of them are $\left \{0.0, 1.0, 3.0, 5.0, 10.0 \right \}$ and $\left \{0.0, 0.25, 0.5, 0.75, 1.0 \right \}$, respectively. For other important hyper-parameters, we find the best value of learning rate from $\left \{0.01, 0.005, 0.001, 0.0005, 0.0001 \right \}$, embedding dimension from $\left \{32, 64, 128, 256, 512 \right \}$, number of GNN layers from $\left \{2, 3, 4, 5\right \}$, batch size from $\left \{32, 64, 128, 256, 512 \right \}$ (except for ogbg-molhiv $\left \{64, 128, 256, 512, 1024 \right \}$). Besides, we fix the perturbation bound $\epsilon$, ascent step size $\alpha$ and ascent step $T$ as 0.008, 0.008 and 5 during hyper-parameter fine-tuning. As for the implementation details of transfer learning, we basically follow the pre-training setting of previous works~\citep{GraphCL, GraphLoG}. 

\section{Proof}
\label{appendix: proof}
\subsection{Proof of Theorem 1}
\label{theorem: objective_appendix}
We repeat Theorem 1 as follows. 

\textbf{Theorem 1. }
\textit{Suppose $f(\cdot)$ is a GNN encoder as powerful as 1-WL test. Let $g_{p}(\cdot)$ elicits only the augmentation information from $\mathbf{z}$ meanwhile $g_{c}(\cdot)$ extracts the essential factors of $G$ from $\mathbf{z}_{1}$ and $\mathbf{z}_{2}$. Then we have:}
\begin{equation}
I\left(t_{1}(G) ; \mathbf{z}_{2}^{c}, \mathbf{z}_{2}^{p}\right) \geq I\left( \mathbf{z}_{1}^{p} ; \mathbf{z}_{2}^{p}\right) \text { where } G \in \mathcal{G} \text{ and } t_{1}(\cdot), t_{2}(\cdot) \in \mathcal{T}.
\nonumber
\end{equation}

\textbf{Proof.} According to the assumption in Theorem \ref{theorem: objective}, for any two graphs $G, G^{\prime} \in \mathcal{G}$, if $G \cong G^{\prime}$ then we have $\mathbf{z}=\mathbf{z^{\prime}}$, where $\mathbf{z}=f(G)$ and $\mathbf{z^{\prime}}=f(G^{\prime})$. 

Besides, $\mathbf{z}^{p}=g_{p}(\mathbf{z})$ is specific to the predictive factors and $\mathbf{z}^{c}=g_{c}(\mathbf{z})$ is particular to the non-predictive factors, which means $\mathbf{z}^{p}$ and $\mathbf{z}^{c}$ are mutually excluded and $\mathbf{z}^{p} \sim G$. So we have, 
% are learned from the same augmented graph view $t(G)$ and they are disentangled (mutually excluded). thus they are independent and conditional independent:  
\begin{equation}
\begin{gathered}
    p\left(\mathbf{z}^{p}, \mathbf{z}^{c}\right) = p\left(\mathbf{z}^{p}\right)p\left(\mathbf{z}^{c}\right) \\
    p\left(\mathbf{z}^{p}, \mathbf{z}^{c} \mid t(G) \right) = p\left(\mathbf{z}^{p} \mid t(G) \right)p\left(\mathbf{z}^{c} \mid t(G) \right) .
\end{gathered}
\label{eq: independent}
\end{equation}

Then, we want to prove that given three random variables $a$, $b$ and $c$, if they satisfy  $p\left( b, c \right)=p\left( b \right)p\left( c \right)$ and $p\left( b, c \mid a \right)=p\left( b \mid a \right)p\left( c \mid a \right)$, we have $I\left( a, b \mid c \right)=I\left(a, b \right)$. According to the definition of mutual information, we have that, 
\begin{equation}
\begin{aligned}
      & I\left(a ; b \mid c\right) = \\ 
      & = \sum_{a} \sum_{b} \sum_{c} p\left(a, b, c\right) \log \frac{p\left(a, b, c\right) p\left(c\right)}{p\left(a, c\right) p\left(b, c\right)}  \\
      & = \sum_{a} \sum_{b} \sum_{c}
      p\left(a\right)p\left(b, c \mid a\right) \log \frac{p\left(b, c \mid a \right) p\left(c\right)}{p\left(c \mid a\right) p\left(b\right)p\left(c\right)} \\ 
      & = \sum_{a} \sum_{b} \sum_{c}
      p\left(a\right)p\left(b \mid a\right) p\left(c \mid a\right) \log \frac{p\left(b \mid a \right)p\left(c \mid a \right)}{p\left(c \mid a\right) p\left(b\right)} \\ & = \sum_{a} \sum_{b}
      p\left(a\right)p\left(b \mid a\right) \log \frac{p\left(b \mid a \right)}{p\left(b\right)} \\ 
      & = \sum_{a} \sum_{b}
      p\left(a, b\right) \log \frac{p\left(b \mid a \right)}{p\left(b\right)} \\ 
      & = I\left(a; b\right) .
\end{aligned}
\label{eq: disen_propo}
\end{equation}

After that, by applying the chain rule to $I\left(t_{1}(G) ; \mathbf{z}_{2}^{p}, \mathbf{z}_{2}^{c}\right)$, we have,
\begin{equation}
\begin{aligned}
    I\left(t_{1}(G) ; \mathbf{z}_{2}^{p}, \mathbf{z}_{2}^{c}\right) 
    & = I\left(t_{1}(G) ; \mathbf{z}_{2}^{p} \mid \mathbf{z}_{2}^{c}\right) + I\left(t_{1}(G) ; \mathbf{z}_{2}^{c}\right) \\
    & \stackrel{(2)}{=} I\left(t_{1}(G) ; \mathbf{z}_{2}^{p} \right) + I\left(t_{1}(G) ; \mathbf{z}_{2}^{c}\right) \\ 
    & \stackrel{(a)}{\geq} I\left(t_{1}(G) ; \mathbf{z}_{2}^{p} \right) \\ 
    & \stackrel{(b)}{\geq} I\left(\mathbf{z}_{1}^{c}, \mathbf{z}_{1}^{p} ; \mathbf{z}_{2}^{p} \right) \\
    & \stackrel{(2)}{=} I\left(\mathbf{z}_{1}^{c}; \mathbf{z}_{2}^{p} \right) + I\left(\mathbf{z}_{1}^{p} ; \mathbf{z}_{2}^{p} \right) \\ 
    & \stackrel{(a)}{\geq} I\left(\mathbf{z}_{1}^{p} ; \mathbf{z}_{2}^{p} \right) ,
\end{aligned}
\end{equation}
where $\stackrel{(2)}{=}$ is derived from the conclusion we get in Equation \ref{eq: disen_propo}, $\stackrel{(a)}{\geq}$ is based on the non-negativity of mutual information, i.e., $I(;) \geq 0$, and $\stackrel{(b)}{\geq}$ is because data processing inequality \citep{DPI}. Finally, we reach to the lower bound of $I\left(t_{1}(G) ; \mathbf{z}_{2}^{p}, \mathbf{z}_{2}^{c}\right)$ in Equation \ref{eq: disen_propo}, thus we can maximize the consistency between the information we capture from the two augmentation graph views by minimizing $\mathcal{L}_{\text{pre}}$. 

\subsection{Proof of Theorem 2}
\label{theorem: disentangle_appendix}
We repeat Theorem 2 as follows. 

\textbf{Theorem 2. }
\textit{Assume $q$ is a Gaussian distribution, $g_{r}$ is the parameterized reconstruction model which infer $\mathbf{z}_{w}$ from $\left( \mathbf{z}_{w^{\prime}}^{p}, \mathbf{z}_{w}^{c} \right)$. Then we have: }
\begin{equation}
H\left( \mathbf{z}_{w} \mid \mathbf{z}_{w^{\prime}}^{p}, \mathbf{z}_{w}^{c} \right) \leq \left\|\mathbf{z}_{w}-  g_{r}\left(\mathbf{z}_{w^{\prime}}^{p} \odot \mathbf{z}_{w}^{c}\right) \right\|_{2}^{2} \text { where } w=w^{\prime} \text{ or } w \neq w^{\prime}.
\nonumber
\end{equation}

\textbf{Proof.} To reconstruct the entangled representation $\mathbf{z}_{w}$ from its corresponding non-predictive representation $\mathbf{z}_{w}^{c}$ and the predictive representation of any augmentation view $\mathbf{z}_{w^{\prime}}^{p}$ ($w$ and $w^{\prime}$ are not necessarily equal), we need to minimize the conditional entropy:
\begin{equation}
    H\left(\mathbf{z}_{w} \mid \mathbf{z}_{w^{\prime}}^{p}, \mathbf{z}_{w}^{c}\right)=-\mathbb{E}_{p\left(  \mathbf{z}_{w}, \mathbf{z}_{w^{\prime}}^{p}, \mathbf{z}_{w}^{c} \right)}\left[\log p\left(\mathbf{z}_{w} \mid \mathbf{z}_{w^{\prime}}^{p}, \mathbf{z}_{w}^{c}\right)\right].
\end{equation}
Since the real distribution of $p\left(\mathbf{z}_{w} \mid \mathbf{z}_{w^{\prime}}^{p}, \mathbf{z}_{w^{\prime}}^{c}\right)$ is unknown and intractable, we hereby introduce a variational distribution $q\left(\mathbf{z}_{w} \mid \mathbf{z}_{w^{\prime}}^{p}, \mathbf{z}_{w}^{c}\right)$ to approximate it. Therefore, we have,
\begin{equation}
\begin{aligned}
    \mathbb{E}_{p\left(  \mathbf{z}_{w}, \mathbf{z}_{w^{\prime}}^{p}, \mathbf{z}_{w}^{c} \right)} & \left[\log p\left(\mathbf{z}_{w} \mid \mathbf{z}_{w^{\prime}}^{p}, \mathbf{z}_{w}^{c} \right)\right]  = \\
    & \mathbb{E}_{p\left(  \mathbf{z}_{w}, \mathbf{z}_{w^{\prime}}^{p}, \mathbf{z}_{w}^{c} \right)}\left[\log q\left(\mathbf{z}_{w} \mid \mathbf{z}_{w^{\prime}}^{p}, \mathbf{z}_{w}^{c}\right)\right] \\
    & + D_{\mathrm{KL}}\left(p\left(\mathbf{z}_{w} \mid \mathbf{z}_{w^{\prime}}^{p}, \mathbf{z}_{w}^{c}\right) \| q\left(\mathbf{z}_{w} \mid \mathbf{z}_{w^{\prime}}^{p}, \mathbf{z}_{w}^{c}\right)\right) .
\end{aligned}
\end{equation}
Due to the non-negativity of KL-divergence between any two distributions, it is safe to say $-\mathbb{E}_{p\left(  \mathbf{z}_{w}, \mathbf{z}_{w^{\prime}}^{p}, \mathbf{z}_{w}^{c} \right)}\left[\log q\left(\mathbf{z}_{w} \mid \mathbf{z}_{w^{\prime}}^{p}, \mathbf{z}_{w}^{c}\right)\right]$ is the upper bound of $H\left(\mathbf{z}_{w} \mid \mathbf{z}_{w^{\prime}}^{p}, \mathbf{z}_{w}^{c}\right)$. Based on the assumption of Theorem \ref{theorem: disentangle}, let $q\left(\mathbf{z}_{w} \mid \mathbf{z}_{w^{\prime}}^{p}, \mathbf{z}_{w}^{c}\right)$ being a Gaussian distribution $\mathcal{N}\left(\mathbf{z}_{w} \mid g_{r}\left(\mathbf{z}_{w^{\prime}}^{p} \odot \mathbf{z}_{w}^{c}\right), \sigma^2 \mathbf{I}\right)$, where $g_{r}(\cdot)$ is the reconstruct network that predict $\mathbf{z}_{w}$ from $\left( \mathbf{z}_{w^{\prime}}^{p}, \mathbf{z}_{w}^{c} \right)$ and $\sigma$ is the variance. Thus we have, 
\begin{equation}
\begin{aligned}
    H\left(\mathbf{z}_{w} \mid \mathbf{z}_{w^{\prime}}^{p}, \mathbf{z}_{w}^{c}\right) 
    & \leq -\mathbb{E}_{p\left(  \mathbf{z}_{w}, \mathbf{z}_{w^{\prime}}^{p}, \mathbf{z}_{w}^{c} \right)}\left[\log q\left(\mathbf{z}_{w} \mid \mathbf{z}_{w^{\prime}}^{p}, \mathbf{z}_{w}^{c}\right)\right] \\
    & = - \mathbb{E}_{p\left(  \mathbf{z}_{w}, \mathbf{z}_{w^{\prime}}^{p}, \mathbf{z}_{w}^{c} \right)}\left[\log \left(\frac{1}{\sqrt{2 \pi I} \sigma }  e^{ -\frac{1}{2} \frac{ \left( \mathbf{z}_{w}-g_{r}\left(\mathbf{z}_{w^{\prime}}^{p} \odot \mathbf{z}_{w}^{c}\right) \right)^{2}}{(\sigma^{2} \mathbf{I})}} \right) \right] \\ 
    & = - \mathbb{E}_{p\left(  \mathbf{z}_{w}, \mathbf{z}_{w^{\prime}}^{p}, \mathbf{z}_{w}^{c} \right)}\left[ \log \left( \frac{1}{\sqrt{2 \pi I} \sigma } \right) -\frac{\left(\mathbf{z}_{w}-g_{r}\left(\mathbf{z}^{p}_{w^{\prime}} \odot \mathbf{z}^{c}_{w}\right)\right)^{2}}{2 \sigma^{2} \mathbf{I}} \right].
\end{aligned}
\label{eq: disen_upper}
\end{equation}
Hence, we get the upper bound of  $H\left(\mathbf{z}_{w} \mid \mathbf{z}_{w^{\prime}}^{p}, \mathbf{z}_{w}^{c}\right)$ as Equation \ref{eq: disen_upper}. To minimize the value of the unsolvable entropy, we can instead minimize the value of its upper bound and thereby derive the objective function as follow by neglecting the constant terms,
% When we remove the constant term $\log \frac{1}{\sqrt{2 \pi \sigma I}}$ and the scaling factor $2 \sigma \mathbf{I}$, we can derive the the optimization function as follow: 
\begin{equation}
\min \mathbb{E}_{p\left(  \mathbf{z}_{w}, \mathbf{z}_{w^{\prime}}^{p}, \mathbf{z}_{w}^{c} \right)}\left\|\mathbf{z}_{w}-g_{r}\left(\mathbf{z}^{p}_{w^{\prime}} \odot \mathbf{z}^{c}_{w} \right)\right\|_{2}^{2} .
\end{equation}
Since we adopt two augmentation views and propose the cross-view reconstruction mechanism in our method, we can minimize the entropy by minimizing $\mathcal{L}_{\text{recon}}$ and thus guarantee the disentanglement of $\mathbf{z}^{p}$ and $\mathbf{z}^{c}$.

\section{Effects of Representation Disentanglement}
\label{appendix: disentangle}
\begin{figure*}[h]
% \vspace{-0.3in}
  \centering
  \includegraphics[width=\linewidth]{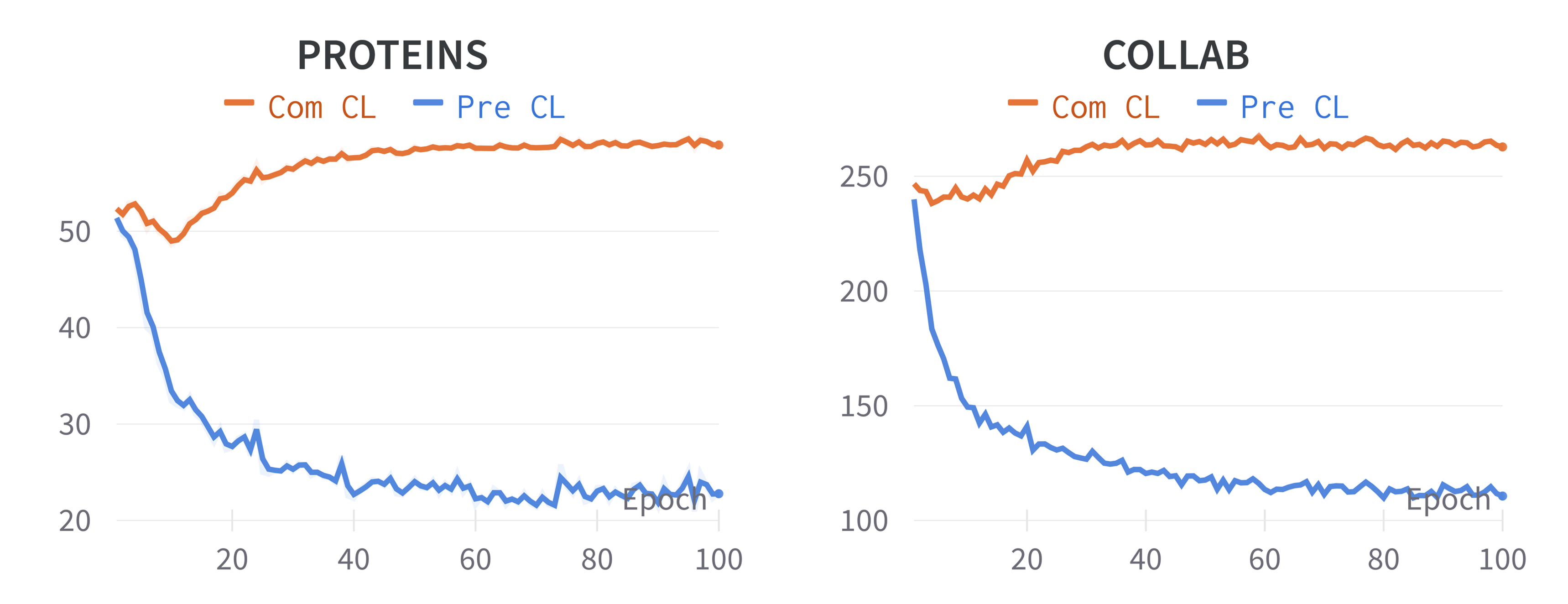}
  \caption{InfoNCE loss of the two disentangled representations between the two augmentation graph views, where orange lines are the InfoNCE loss between the two non-predictive representations and blue lines are the InfoNCE loss between the two predictive representaitons} 
  \vspace{-0.2in}
  \label{fig: CL_Loss}
\end{figure*}

In this section, we set experiments to investigate the representation disentanglement of our proposed GraphCV. Specifically, we use the InfoNCE loss \citep{InfoNCE} to dynamically measure the representation difference between the two augmentation graph views based on the two disentangled representations, where blue lines indicate the InfoNCE loss between $\mathbf{z}^{p}_{1}$ and $\mathbf{z}^{p}_{2}$ and orange lines represent the InfoNCE loss between $\mathbf{z}^{c}_{1}$ and $\mathbf{z}^{c}_{2}$. For simplicity, we only demonstrate the first 100 pre-training epochs of PROTEINS and COLLAB in Figure \ref{fig: CL_Loss}, we can observe similar phenomena on other datasets. From the loss curves in Figure \ref{fig: CL_Loss} we can find that contrastive loss between predictive representations gradually decreases, indicating the predictive representation is optimized to capture all the shared information between the two augmentation view. Meanwhile, we can see contrastive loss between the non-predictive representations achieve a noticeable increases, which is consistent with our expectation that the two independent sampled augmentation operators cause a distribution shift between the two augmentation views.  To further investigate whether the feature suppression problem is equally serious in $\mathbf{z}^{p}$ and $\mathbf{z}^{c}$, we conduct experiments to compare the performance of the two representation on downstream tasks. The comparison results are as follow:

\begingroup
\begin{table*}[h]
\centering
\caption{Performance comparison of the two learned representations. Results are reported as mean±std\%, the best performance is bolded.}
\label{tab: disen}
\setlength{\tabcolsep}{3pt}
\begin{adjustbox}{width=\textwidth,center}
\begin{tabular}{ccccccccccc} 
\toprule
  & \textbf{MUTAG} & \textbf{COLLAB} & \textbf{NCI1} & \textbf{PROTEINS}  & \textbf{IMDB-B}   & \textbf{RDT-B}  & \textbf{DD} & \textbf{ogbg-molhiv}\\
\midrule
$\mathbf{z}^{c}$ &88.1±1.2    & 75.1±0.7&	72.2±2.0&	73.5±0.8&	71.8±0.9&	89.4±1.0&	75.8±0.6&	69.7.0±2.8\\
$\mathbf{z}^{p}$&	\textbf{92.3±0.7}&	\textbf{80.5±0.5}&	\textbf{82.0±1.0}&	\textbf{76.8±0.4}&	\textbf{75.6±0.4}&	\textbf{92.5±0.9}&	\textbf{80.5±0.5} &\textbf{75.36±1.4}\\
\bottomrule
\end{tabular}
\end{adjustbox}
\end{table*}
\endgroup

It is easily to observe that there is a obvious performance gap between the two learned representation, indicating the different feature suppression issue between them and the features subset that are more robust to augmentation is more informative and transferable that those sensitive to augmentations. Therefore, we believe our proposed GraphCV can further alleviate the feature suppression issue with the disentanglement design.   

\section{Training Algorithm}
\label{appendix: alg}
In this sectionm we summarized the details of our proposed method in the following Algorithm.
\begin{algorithm}[h]
\caption{The training algorithm of \textbf{GraphCV}}\label{alg:HDGAT}
    \SetKwInOut{Input}{Input}
    \SetKwInOut{Output}{Output}

    \KwIn{Graph dataset $\mathcal{G}=\left\{G_{i} = (V_{i}, E_{i})\right\}_{i=1}^{N}$; augmentation family $\mathcal{T}$; loss coefficient $\lambda_{r}$, $\lambda_{a}$; ascernt step $T$; ascent step size $\alpha$; perturbation bound $\epsilon$. }
    \KwOut{The disentangled predictive representations $\mathbf{Z}^{p}=\left\{ \mathbf{z}_{i}^{p} \right\}_{i=1}^{N}$}
    \For{each training epoch}{
        \For{sampled minibatch $\mathcal{B}=\left\{G_{i}\right\}_{i=1}^{|\mathcal{B}|}$}{
        \For{$G_{i} \in \mathcal{B}$} {
        { $\mathbf{z}_{1, i}=f\left(t_{1}(G_{i})\right)$, $\mathbf{z}_{2, i}=f\left(t_{2}(G_{i})\right)$ ; \Comment{$t_{1}(\cdot), t_{2}(\cdot) \in \mathcal{T}$} \\
        $\mathbf{z}_{1, i}^{p}=g_{p}\left( \mathbf{z}_{1, i} \right)$, $\mathbf{z}_{2, i}^{p}=g_{p}\left( \mathbf{z}_{1, i} \right)$ ; \\
        $\mathbf{z}_{1, i}^{c}=g_{c}\left( \mathbf{z}_{1, i} \right)$, $\mathbf{z}_{2, i}^{c}=g_{c}\left( \mathbf{z}_{1, i} \right)$ ;}
        }
        Calculate $\mathcal{L}_{\text{pre}}$ according to Equation 6; \\
        Calculate $\mathcal{L}_{\text{recon}}$ according to Equation 8 ; \\
        $\mathcal{L} \leftarrow \mathcal{L}_{\text{pre}} + \lambda_{r} \mathcal{L}_{\text{recon}}$; \\
        $\delta_{0} \leftarrow U(-\epsilon, \epsilon)$; \\
        
        \For{each $t=1$ to $T$}{
        { Calculate the $\mathcal{L}_{\text{adv}}$ according to Equation 10; \\
        $\delta_{t} \leftarrow \delta_{t-1} + \alpha\nabla_{\delta} \mathcal{L}_{\text{adv}}$; \Comment{Update perturbation to maximize $\mathcal{L}_{\text{adv}}$}\\ 
        $\mathcal{L} \leftarrow \mathcal{L} + \frac{\lambda_{a}}{T} \mathcal{L}_{\text{adv}}$ 
        }
        }
        Update the parameter $\theta$ of $f$ and $g$ with the gradient $\nabla_{\theta} \mathcal{L}(\theta, \mathcal{B})$ over a minibatch; 
    }
    }
    \Return $\mathbf{Z}^{p}=\left\{ \mathbf{z}_{i}^{p} \right\}_{i=1}^{N}$, where $\mathbf{z}_{i}^{p} = g_{p}\left( f(G_{i}) \right)$\\
\end{algorithm}

\section{Hyper-parameter Sensitivity}
\label{sec: appendix_hyper}
In this section, we study the impacts of some important hyper-parameters in our method, including reconstruction loss coefficient $\lambda_{r}$, adversarial loss coefficient $\lambda_{a}$, embedding dimension $d$, batch size $|\mathcal{B}|$ and number of GNN layers $L$. Here, we select four datasets, i.e., MUTAG, PROTEINS, RDT-B and COLLAB, to report for simplicity because the four datasets cover different domains and scales. We illustrate the impacts of these hyper-parameters in the figures below. 

\begin{figure*}[h!]
  \centering
  \includegraphics[width=\linewidth]{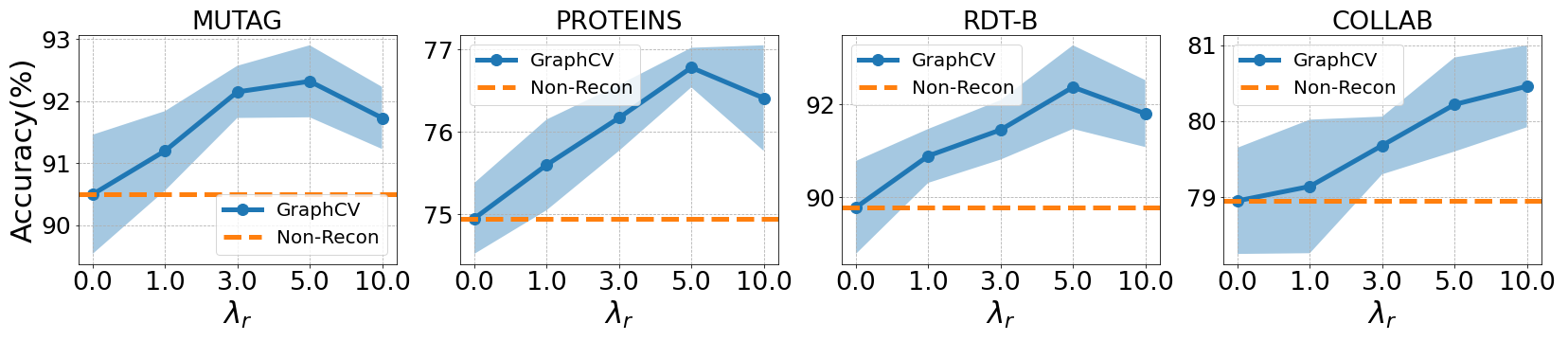}
  \caption{Impact of reconstruction loss coefficient $\lambda_{r}$ on different datasets, we specify the non-reconstruction situation ($\lambda_{r}=0$) with the dashed line for comparison.} 
  \label{fig: lamr}
\end{figure*}

From the result demonstrated in Figure \ref{fig: lamr}, we can see 
% our proposed cross-view reconstruction mechanism can obviously alleviate the entangled issue we mentioned and improve the performances. 
the optimal reconstruction loss coefficient $\lambda_{r}$ is different dependent on the specific dataset, but all the values in our experiment can enhance the performance compared with non-reconstruction variant, i.e., $\lambda_{r}=0$, indicating the effectiveness of our proposed cross-view reconstruction mechanism. 

\begin{figure*}[h!]
  \centering
  \includegraphics[width=\linewidth]{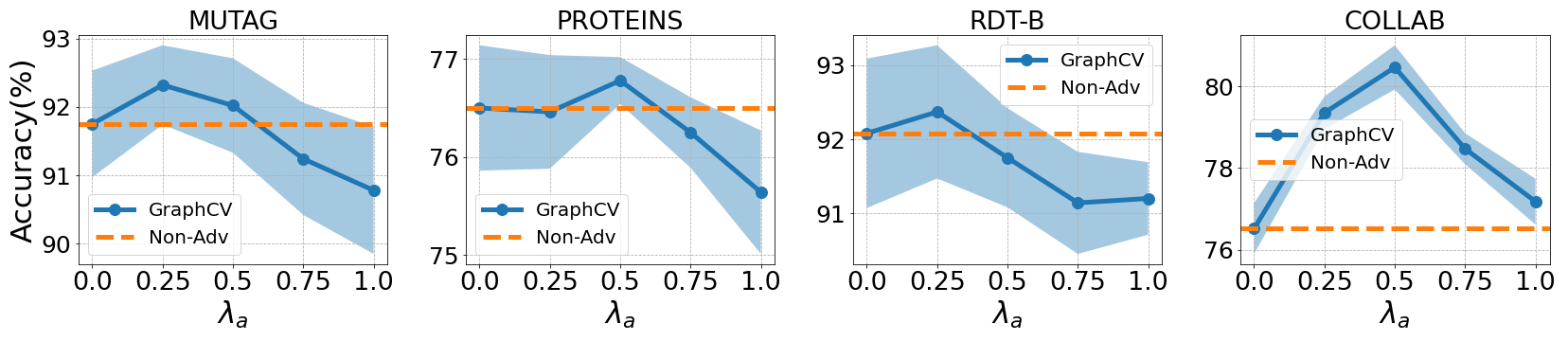}
  \caption{Impact of adversarial loss coefficient $\lambda_{a}$ on different datasets, we specify the non-adversarial situation ($\lambda_{a}=0$) with the dashed line for comparison.}
  \label{fig: lama}
\end{figure*}
The Figure \ref{fig: lama} shows that we could further raise the model performance through the adversarial training, which proves a robust representation with less redundant information usually achieve more performance gain compared with the brittle one. During this process, we need to choose a appropriate adversarial loss coefficient $\lambda_{a}$, otherwise a too large $\lambda_{a}$ may hurt the information sufficiency of the learned representation.

\begin{figure*}[h!]
  \centering
  \includegraphics[width=\linewidth]{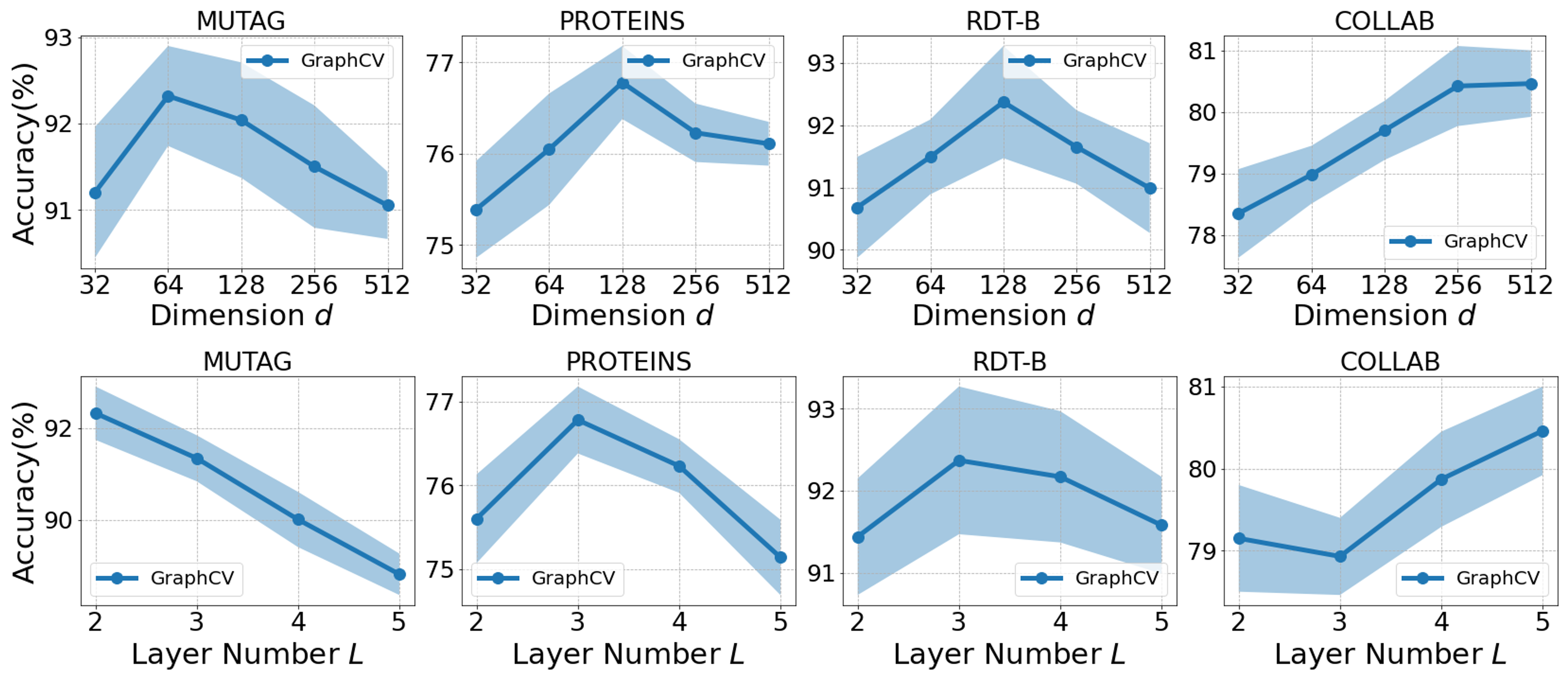}
  \caption{Impact of a embedding dimension $d$ and GNN layer number $L$ on different datasets.} 
  \label{fig: emb_lay}
\end{figure*}

We put the impacts of embedding dimension $d$ and GNN layer number $L$ together because we can find a similar observation form their experimental results. From Figure \ref{fig: emb_lay}, we observe that the optimal values of the two hyper-parameters generally increase as the dataset scale increases. The reason behind this phenomena could be large datasets usually contain more latent factors than the small datasets, therefore a model with larger capacity is needed to fit the large datasets. However, such high-capacity message-passing model will deteriorate the performance of small dataset because it may cause the learned representation over-smoothing and hence less informative. 

% \begin{figure*}[h!]
%   \centering
%   \includegraphics[width=\linewidth]{bs.png}
%   \caption{Impact of batch size $|\mathcal{B}|$ on different datasets.} 
%   \label{fig: bs}
% \end{figure*}

% The effect of batch size $|\mathcal{B}|$ is shown in Figure \ref{fig: bs}, we can see the performance variation of these datasets under different batch size is relatively small. Most of the datasets achieve best performance when the $|\mathcal{B}|$ is set to 128. 

\end{document}